\documentclass{article}
\usepackage{ijcai23}

\usepackage{times}
\usepackage{soul}
\usepackage{url}
\usepackage[hidelinks]{hyperref}
\usepackage[utf8]{inputenc}
\usepackage[small]{caption}
\usepackage{graphicx}
\usepackage{amsmath}
\usepackage{amsthm}
\usepackage{booktabs}
\usepackage{tabularx}
\usepackage{multirow}
\usepackage{color}
\usepackage[utf8]{inputenc}
\usepackage{booktabs}
\usepackage[table]{xcolor}
\usepackage{siunitx}
\usepackage{multirow}

\usepackage{algorithm,algorithmic}
\usepackage{amssymb}
\usepackage{amsfonts}
\usepackage{amsthm}
\usepackage{mathtools}
\usepackage{bbding}
\usepackage{multicol}
\usepackage{multirow}
\usepackage{scalerel}
\usepackage{subcaption}
\usepackage{color}
\usepackage{xcolor}
\usepackage[toc,page]{appendix}
\usepackage{colortbl}
\usepackage{diagbox}
\usepackage{stfloats}
\usepackage{xpatch}
\usepackage{tablefootnote}

\newcommand{\Fref}[1]{Figure~\ref{#1}}
\newcommand{\Sref}[1]{Sec.~\ref{#1}}
\newcommand{\Aref}[1]{Alg.~\ref{#1}}

\makeatletter
\newif\ifdiagbox@cellEmpty@

\define@key{diagbox}{widest}{%
  \ifdiagbox@cellEmpty@
    \setkeys{diagbox}{innerwidth=\widthof{#1}}%
    \PackageWarning{diagbox}{innerwidth is set to \the\diagbox@wd}%
  \fi}

\define@key{diagbox}{highest}{%
  \ifdiagbox@cellEmpty@
    \setkeys{diagbox}{height=#1}%
  \fi}

\define@key{diagbox}{text}{%
  \def\diagbox@text{#1}}

\define@key{diagbox}{align}{%
  \in@{#1}{l, c, r}%
  \ifin@
    \def\diagbox@align{#1}%
  \else
    \PackageError{diagbox}%
      {The value of key "text" must be one of "l", "c", and "r".}%
  \fi}

\xpatchcmd{\diagbox@double}{%
  \setkeys{diagbox}{dir=NW,#1}%
}{%
  \if\relax\detokenize{#2}\relax
    \if\relax\detokenize{#3}\relax
      \diagbox@cellEmpty@true
      \setkeys{diagbox}{highest=1\line, align=l, text=\@empty}%
    \fi
  \fi
  \setkeys{diagbox}{dir=NW, #1}%
  \ifdiagbox@cellEmpty@
    \rlap{\makebox
      [\dimexpr\diagbox@wd-\diagbox@insepl-\diagbox@insepr\relax]%
      [\diagbox@align]%
      {\diagbox@text}}%
  \fi
}{}{\ddt}
\makeatother

\newfloat{figtab}{htb}{fgtb}
\makeatletter
  \newcommand\figcaption{\def\@captype{figure}\caption}
  \newcommand\tabcaption{\def\@captype{table}\caption}
\makeatother

\DeclareMathOperator*{\argmin}{argmin} 
\DeclareMathOperator*{\argmax}{argmax} 

\newtheorem{theorem}{Theorem}

\newtheorem{prop}{Proposition}
\newtheorem{rmk}{Remark}

\newcommand{\calD}{\mathcal{D}}

\newcommand{\calC}{\mathcal{C}}

\newcommand{\RR}{\mathbb{R}}

\newcommand{\gray}[1]{{\color{gray}#1}}

\usepackage{pifont}
\title{Unlearning during Learning: An Efficient Federated Machine Unlearning Method}

\usepackage[switch]{lineno}

\author{
Hanlin Gu$^{1}$
\and
Gongxi Zhu$^{2,3}$\and
Jie Zhang$^{4}$\and
Xinyuan Zhao$^{3}$\and \\
Yuxing Han$^{3}$\and
Lixin Fan$^{1}$\And
Qiang Yang$^{1}$
\affiliations
$^1$AI Lab, Webank \and \\ $^2$ University of Electronic Science Technology of China \and \\ $^3$
Shenzhen International Graduate School, Tsinghua University  \and  \\
$^4$ Nanyang Technological University 
\emails
allengu@webank.com, gx.zhu@foxmail.com, jie\_zhang@ntu.edu.sg, 
yuxinghan@sz.tsinghua.edu.cn
}

\begin{document}

\maketitle

\begin{abstract}
In recent years, Federated Learning (FL) has garnered significant attention as a distributed machine learning paradigm. 
To facilitate the implementation of the ``right to be forgotten," the concept of federated machine unlearning (FMU) has also emerged. However, current FMU approaches often involve additional time-consuming steps and may not offer comprehensive unlearning capabilities, which renders them less practical in real FL scenarios.
In this paper, we introduce FedAU, an innovative and efficient FMU framework aimed at overcoming these limitations. Specifically, FedAU incorporates a lightweight auxiliary unlearning module into the learning process and employs a straightforward linear operation to facilitate unlearning. This approach eliminates the requirement for extra time-consuming steps, rendering it well-suited for FL.
Furthermore, FedAU exhibits remarkable versatility. It not only enables multiple clients to carry out unlearning tasks concurrently but also supports unlearning at various levels of granularity, including individual data samples, specific classes, and even at the client level.
We conducted extensive experiments on MNIST, CIFAR10, and CIFAR100 datasets to evaluate the performance of FedAU. The results demonstrate that FedAU effectively achieves the desired unlearning effect while maintaining model accuracy. Our code is availiable at \url{https://github.com/Liar-Mask/FedAU}.

\end{abstract}

\section{Introduction}
Federated learning (FL) \cite{konevcny2015federated,mcmahan2017communication,yang2019federated} is a promising distributed machine learning paradigm that provides privacy-preserving learning solutions. One essential requirement of FL is the participants’ “right to be forgotten”, which has been stated
explicitly in the European Union General Data Protection
Regulation (GDPR)\footnote{\url{https://gdpr-info.eu/art-17-gdpr/}} and the California Consumer
Privacy Act (CCPA) \cite{harding2019understanding}.
Federated Machine Unlearning (FMU) is proposed to give clients the right to remove the influence of a certain subset of their data from
a trained federated learning (FL) model, while maintaining the accuracy of the FL model on remaining data \cite{che2023fast}.

Three representative existing FMU approaches have been proposed. Firstly, One prevalent approach involves the retraining or fine-tuning of the model from scratch using the \textit{remaining data}\cite{liu2022right,liu2021federaser,su2023asynchronous,zhang2023fedrecovery}.  Secondly, another line of research explores the utilization of Gradient Ascent on the \textit{unlearning data} to effectively diminish its impact \cite{wu2022federated,graves2021amnesiac}. Thirdly, \cite{wang2022federated} explored the application of model pruning techniques. Specifically, they selectively removed certain neurons from the model architecture that exhibit a high correlation with the unlearning data. In practice, there are two important ingredients required for FMU \cite{zhang2023fedrecovery,liu2023survey}:
\begin{itemize}
\item \textbf{Reduced Unlearning Time}: FL systems require FMU methods that minimize the time required for unlearning operations. This is crucial because normal clients participating in FL cannot afford to wait for the unlearning client to complete the unlearning process. Even for methods like gradient ascent \cite{wu2022federated,graves2021amnesiac} and pruning \cite{wang2022federated}, there is still a need for a certain amount of time to implement the unlearning operation. 

\item \textbf{Broad Unlearning Capability}: An effective FMU method should have the capability to accommodate unlearning requests from multiple clients in FL. This includes the ability to unlearn specific samples, classes, or clients as requested by different clients participating in the FL process. 
\end{itemize}

However, existing methods do not consider these two important requirements simultaneously. In order to satisfy these two requirements, we propose an efficient Federated Machine Unlearning (FMU) framework called FedAU. FedAU incorporates an auxiliary unlearning module during the training that facilitates the unlearning process. Our framework offers three key advantages:
Firstly, FedAU utilizes a simple linear operation to achieve unlearning, which avoids consuming the waiting time for other normal clients during the federated learning process (see Sect. \ref{subsec:fedau}). Secondly, FedAU is a general unlearning framework that allows multiple clients to implement unlearning. It supports unlearning at the sample, class, and client levels, providing flexibility in managing privacy concerns (see Sect. \ref{subsec:algorithm}).
Thirdly, the proposed FedAU demonstrates strong performance in terms of unlearning effectiveness and model accuracy. This is supported by both theoretical analysis and experimental evaluations (see Sect. \ref{subsec:theo} and Sect. \ref{sec:exp}).

\noindent\textbf{Contribution.}  The main contributions are summarized as follows:
\begin{itemize}
    \item We point out that existing methods for federated machine unlearning (FMU) is not feasible in practice, in terms of unlearning time and unlearning capability.

    \item In this paper, we propose FedAU, 
a streamlined FMU framework, that incorporates a lightweight auxiliary unlearning module into the learning process and adopts a linear operation to achieve unlearning.

    \item Extensive experiments and theoretical analysis demonstrated that FedAU is highly effective in enabling unlearning across various scenarios.

\end{itemize}



\section{Relate Work}
\subsection{Federated Learning}

Federated learning \cite{konevcny2015federated,mcmahan2017communication,cheng2020federated} aims to build a machine learning model based on datasets that are distributed across multiple devices without sharing private data with the server and other devices.
The cornerstone of federated learning algorithms, FedAvg, proposed by~\cite{mcmahan2017communication}, involves local clients training models on their data and sending model updates to a central server. The server then averages these updates to improve a global model. 
Afterward, researchers have proposed various optimization techniques~\cite{deng2020distributionally,sun2022decentralized} to enhance FedAvg.

Given the privacy-centric nature of federated learning, a plethora of research focuses on enhancing data privacy. Techniques such as differential privacy \cite{dwork2006differential} and secure multi-party computation \cite{goldreich1998secure} are often integrated into federated learning algorithms to protect client data.
Nevertheless, recent research has highlighted vulnerabilities in federated learning to privacy breaches, notably through model inversion attacks  \cite{nasr2019comprehensive} and membership inference attacks \cite{he2019model}. In this paper, we focus on leveraging unlearning techniques to mitigate these privacy risks in federated learning scenarios. Notably, we utilize FedAvg as the default federated learning algorithm.

\subsection{Machine Unlearning}

Machine unlearning \cite{bourtoule2021machine,mercuri2022introduction} involves removing the influence of specific training data from a machine learning model, often for privacy, fairness, or data quality reasons. It is a response to challenges like the ``right to be forgotten" in the context of data privacy regulations.
A pivotal advancement in machine unlearning is the development of a unified PAC-Bayesian framework \cite{Jose2021A}, which recasts variational unlearning and forgetting Lagrangian as information risk minimization problems. Another significant development is the introduction of cryptographic frameworks for verifiable machine unlearning \cite{Eisenhofer2022Verifiable}, which allows users to verify the removal of their data.

The application of machine unlearning in federated learning environments presents unique challenges and opportunities \cite{Liu2022The}. 
%
Traditional methods, such as retraining models on remaining data \cite{liu2022right,su2023asynchronous} or directly modifying the original model \cite{liu2021federaser,halimi2022federated,wu2022federated}, are often too time-intensive to be viable in the dynamic setting of federated learning. Furthermore, the federated learning paradigm involves multiple clients, each potentially requesting data unlearning. This scenario adds complexity, as current methodologies rarely address the efficient unlearning of data from numerous clients simultaneously. Additionally, while some methods utilize noise addition for efficiency \cite{sekhari2021remember,zhang2023fedrecovery}, this approach can compromise the performance of models trained on the remaining data, leading to a trade-off between efficiency and model accuracy.
In this paper, our goal is to develop a streamlined approach to federated machine unlearning, adaptable across a range of applications.

\section{The Proposed Method} \label{sec:method}
\begin{figure*}[!ht]
\centering
\includegraphics[width=0.98\textwidth]{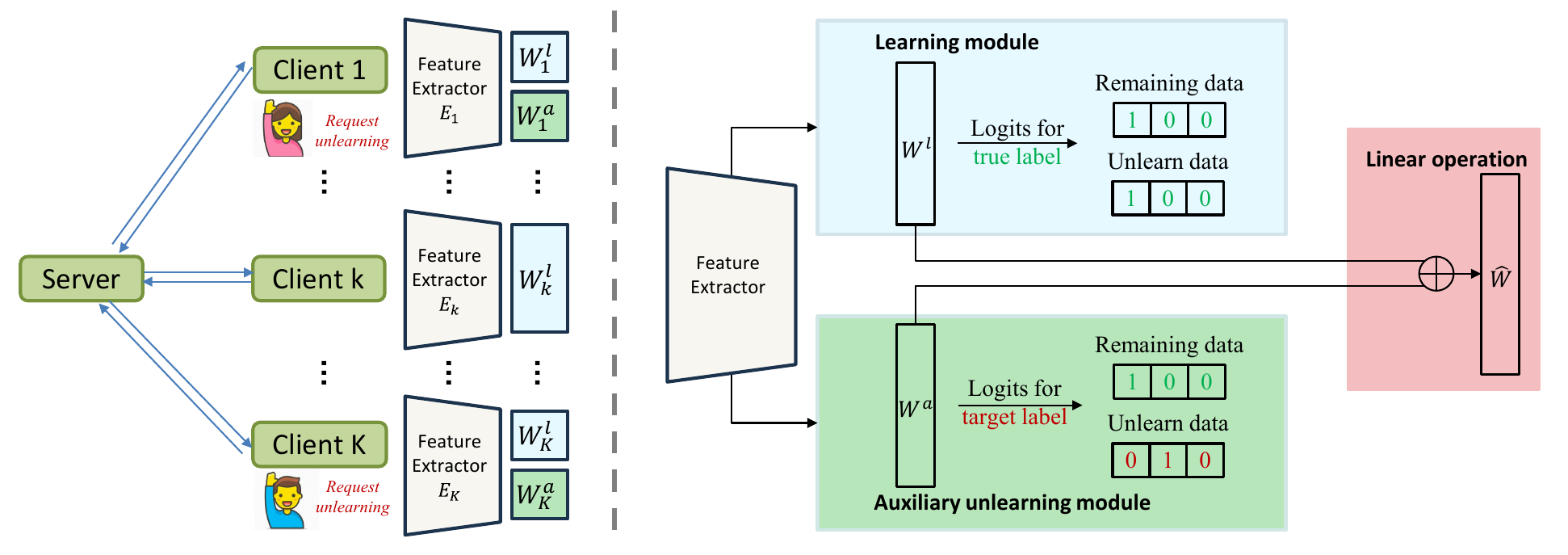}
 \vspace{-1em}
\caption{Left: the scenario of federated machine unlearning; Right: the overview of the proposed FedAU consisted of three modules/operations (blue: learning module, green: auxiliary unlearning module and red: linear operation). }
\label{fig:fm}
\vspace{-3pt}
\end{figure*}


We introduce the FMU setting in Sect. \ref{subsec:setting}, followed by elaboration on the proposed FMU framework, FedAU, in Sect. \ref{subsec:fedau}. Then we show the generality of FedAU in Sect. \ref{subsec:algorithm}: 1) it can be applied in unlearning sample, class, and client; 2) it allows multiple clients to request to unlearn. Finally,  we provide a theoretical analysis of the influence of FedAU on model accuracy and the unlearning effect. in Sec. \ref{subsec:theo}.

\subsection{Setting} \label{subsec:setting}
Consider a Horizontal Federated Learning setting consisting of $K$ clients who collaboratively train a FL model $\omega = (E,W)$ (Feature extractor $E$ and Classifier $W$) to optimize the following objective:
\begin{equation} \label{eq:FL-objective}
    \min_{\omega} \sum_{k=1}^K\sum_{i=1}^{n_k}\frac{\ell(F_{E,W}(x_{k,i}), y_{k,i})}{n_1+\cdots+n_K},
\end{equation}
where $\ell$ is the loss, e.g., the cross-entropy loss and $\calD_k=\{(x_{k,i}, y_{k,i})\}_{i=1}^{n_k}$ is the dataset with size $n_k$ owned by client $k$.
Client $k_0$ requests to withdraw their consent for the utilization of its data, resulting in the need for the server to remove the influence of the data contributed by clients $k_0$ from the trained model. Moreover, if multiple clients request to unlearn simultaneously, we define the set of these multiple clients to be $\calC$.

We note three distinct cases within the Federated Model Unlearning (FMU) framework
\begin{enumerate}
    \item \textit{Unlearning samples}: In this case, the goal is to eliminate the knowledge acquired from a subset of client data, thereby excluding it from the global model.
    \item \textit{Unlearning class}: This case involves excluding a specific class from the model's generalization boundary, effectively removing it from the model's predictions.
    \item \textit{Unlearning client}: Here, the objective is to completely erase the data of a particular client, denoted as $\calD_{k_0}^u = D_{k_0}$, ensuring that the model is no longer influenced by any data from that client.  The detailed experiment to analyze the unlearning effect for the the number of unlearning samples $|\calD_{k_0}^u|$ is illustrated in Appendix C. 
\end{enumerate}

\vspace{1em}
\subsection{FedAU} \label{subsec:fedau}
As shown in the right of \Fref{fig:fm}, our main approach involves the incorporation of an auxiliary unlearning module $W^a$ during the training process of the dataset. When a client requests to unlearn specific information, a straightforward linear operation, such as a weighted average, can be taken between the learning module $W^l$ and the auxiliary unlearning module $W^a$ to produce the final unlearning model $\hat{W}$. More details are provided below.

\subsubsection{Learning Module}
Consider the task of supervised classification using Deep Neural Networks (DNNs). Let $\mathcal{Y} = \{1, ..., C\}$ denote the label space, where $C$ represents the total number of classes. The learning module aims to optimize the following objective:

\begin{equation} \label{eq:objective}
E, W^l = \argmin_{E,W} \sum_{k=1}^K\sum_{(x_{k, i},y_{k, i}) \in \mathcal{D}_k} \frac{\ell(F_{E,W}(x_{k, i}),y_{k, i})}{n_1+ \cdots+n_K}.
\end{equation}
Here, $\ell$ represents the loss function, such as the cross-entropy loss, and $\mathcal{D}_k=\{(x_{k,i}, y_{k,i})\}_{i=1}^{n_k}$ represents the dataset of client $k$ with a size of $n_k$.

\subsubsection{Auxiliary Unlearning Module}
We design an auxiliary unlearning module $W^{a}$ that is learned by clients $\calC$ who request to unlearn their data. The goal of the auxiliary unlearning module is to learn a special model $W^{a}_{k_0}$ for the designed data $D_{k_0}^{'}$ of client $k_0$ as:
\begin{equation} \label{eq:aux-objective}
W^{a}_{k_0} = \argmin_W \sum_{(x_{k_0, i},y_{k_0, i}) \in \mathcal D_{k_0}^{'}}  \frac{\ell(F_{E,W}(x_{k_0, i}),y_{k_0, i})}{|\calD'_{k_0}|}.
\end{equation}
Then we can implement the simple linear operation between $W^l$ and $W^{a}_{k_0}$ in the following section to obtain the unlearning model $\hat{W}$, which can remove the influence of the unlearning data $D^u_{k_0}$.
The auxiliary unlearning module has two characteristics: 1) it is trained during the learning process and efficiently converge with the several training epoch when initializing as $W^l$; 2) multiple unlearning clients $\calC$ can train their own auxiliary unlearning privately or collaboratively to deal with different condition (see unlearning sample in \Sref{subsec:algorithm}).

\subsubsection{Linear Operation on $W^l$ and $W^a$}
The unlearning model $\hat{W}$ needs to satisfy two requirements for unlearning data $D^u$ and remaining data $\calD^r = \calD- \calD^u$. The \textit{\textbf{first requirement}} is that unlearning doesn't influence the model accuracy of the remaining data $\calD^r$. Specifically, the logit output of $\hat{W}$ represents the same to the $W^l$ w.r.t the remaining data $\calD^r$, i.e.,
\newcounter{chemeqn}
\newenvironment{chemequations}
{\def\theequation{R\thechemeqn}}{}
\setcounter{chemeqn}{1}
\begin{chemequations} 
\begin{equation} \label{eq:rq1}
        \argmax_iF^i_{E, W^l}(x) = \argmax_iF^i_{E, \hat{W}}(x), x\in \calD^r,
\end{equation}
\end{chemequations}
where $F_{E,W}(x)$ is the logit output with size $C$ by the trained model on the input $x$ and $F_{E,W}^i(x)$ is the $i_{th}$ logit.  
The \textit{\textbf{second requirement}} is that the model after unlearning behaves wrongly the unlearning data $\calD^u$ such as \cite{chen2023boundary}. Specifically, it requires the logit output of $\hat{W}$ shows the difference to the unlearning data $\calD^u$, i.e., 
\setcounter{chemeqn}{2}
\begin{chemequations}
\begin{equation} \label{eq:rq2}
    \argmax_i F^i_{E, \hat{W}}(x) \neq y , x\in \calD^u,
\end{equation}
\end{chemequations}
where $y$ is true label of $x$. In other words, this requirement indicates that the model after unlearning doesn't memorize the unlearning data $\calD^u$.
\begin{rmk}
Some methods \cite{graves2021amnesiac} aims to achieve the requirement \eqref{eq:rq2} by finetuning the
trained model with randomly labeled forgetting data, but this will also shift the boundary of the remaining class randomly, leading to the degeneration of utility the on remaining data.
\end{rmk}

To make the simple linear operation as described by Eq. \eqref{eq:lr-op} and concurrently to satisfy the aforementioned two requirements (\eqref{eq:rq1} and \eqref{eq:rq2}), we leverage the linear property of a fully connected layer to achieve this:
\begin{equation}\label{eq:lr-op}
    \hat{W} = W^l \bigoplus W^a.
\end{equation}
The following proposition illustrates that the change in logits is proportional to the change in weights if the network is fully connected layer:

\begin{prop}
Consider two fully connected layers projecting the input $x \in \RR^{m_2}$ to the logit $l \in \RR^{m_1}$ as: $l_1= w_1x + b_1, l_2 = w_2x +b_2 $, then the linear operation of weights $w_1, b_1$ and $w_2, b_2$ has the same influence on logits $l_1$ and $l_2$.
\end{prop}

Therefore, by utilizing this property, the model change can effectively reflect the change in logits, thereby satisfying the aforementioned requirements. The specific design of this linear operation for unlearning samples and classes are introduced in the following section.
\begin{rmk}
There is no need to train Auxiliary unlearning 
module at the beginning of learning. As indicated in Appendix C, the training of the AU only necessitates a
few epochs. Consequently, clients can proactively train
the AU module several epochs in advance of the unlearning request rather than at the beginning of learning.
\end{rmk}

\begin{figure}[!ht]
    \centering
     \begin{subfigure}{0.49\textwidth}
  		 	  \includegraphics[width = \linewidth]{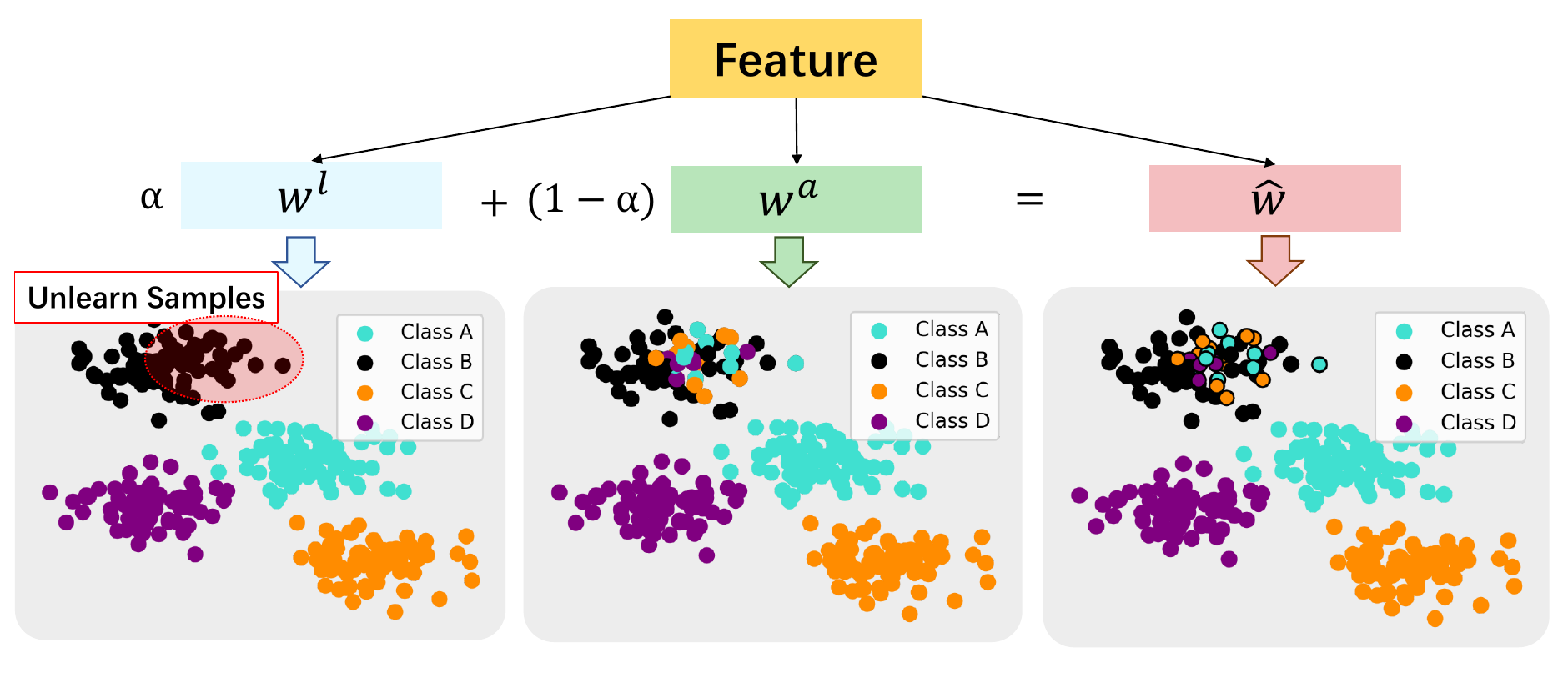}
         \subcaption{Unlearning Sample} 
    	\end{subfigure}

    \begin{subfigure}{0.49\textwidth}
  		 \includegraphics[width=1\textwidth]{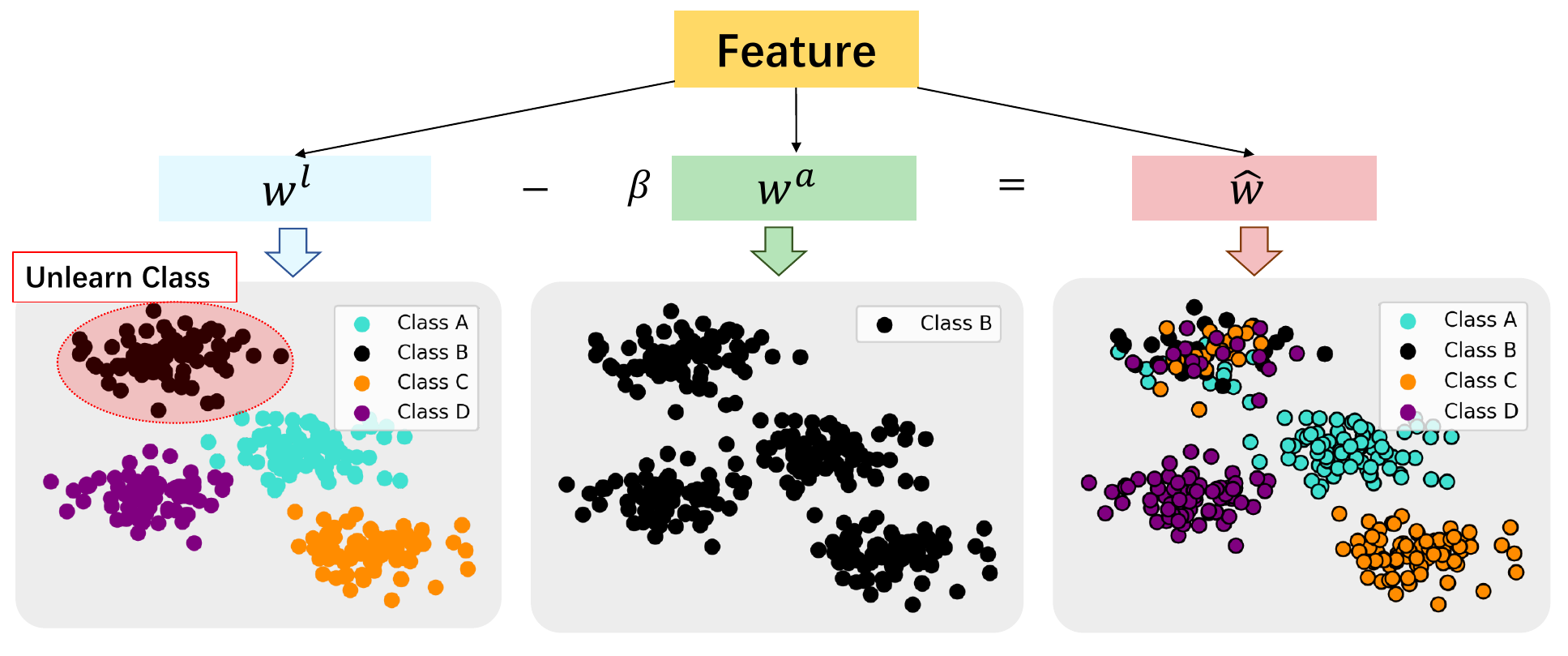}
              \subcaption{Unlearning Class}
    		\end{subfigure}
    \caption{Illustration of the proposed FedAU when unlearning sample and class. After the module $W^l$ undergoes linear operation with the auxiliary unlearning module  $W^a$, the unlearned part of the original feature will be classified into other random classes. }
  
    \label{fig:method}
\end{figure}

\vspace{-4pt}
\subsection{Generality of FedAU} \label{subsec:algorithm}
We provide the details of FedAU on how to unlearn the sample and class in this part (see \Aref{algo:unlearn-sample} and \Aref{algo:unlearn-class}).
\subsubsection{Unlearning Sample in FL}
We firstly consider the scenario that only one client $k_0$ attempts to unlearn some samples $\calD^u_{k_0} = \{(x_{k_0,i}^u, y_{k_0,i}^u)\}_{i=1}^m$, the core steps are as followings:
\begin{itemize}
    \item  Client $k_0$ \textit{designs an the auxiliary dataset} $\calD^{'}_{k_0} = \calD^{u'}_{k_0} \cup \calD_{k_0}^{r'}$. Specifically, $\calD^{u'}_{k_0}$ is based on $\calD^u_{k_0}$ by modifying the label $y_{k,i}^u$ with $y^{u'}_{k,i} \sim U(1,C)$ and $\calD^{r'}_{k_0} =\calD^{r}_{k_0}$, where $U(1,C)$ represents the discrete uniform distribution on value $1,\cdots, y^u_{k_0,i}-1,  y^u_{k_0,i}+1, \cdots,C$  (see blue line 3-9 of Algo. \ref{algo:unlearn-sample}).
    \item Then client $k_0$ \textit{learns the auxiliary unlearning module} $W^a_{k_0}$ according to the $\calD^{'}_{k_0}$ during the learning stage (see green line 10-15 of Algo. \ref{algo:unlearn-sample}).
    \item Finally, when the unlearning request is proposed, the unlearning model $\hat{W}$ can be obtained as:
\begin{equation} \label{eq:lr-sample-unlearn}
    \hat{W}  = \alpha W^l + (1-\alpha) W^a_{k_0},
\end{equation}
where $\alpha$ is a small positive coefficient (see red line 21 of Algo. \ref{algo:unlearn-sample}).
\end{itemize} 
As shown in Fig. \ref{fig:method}(a), the the class of remaining data is not influenced by the addition operation since the remaining and auxiliary dataset have the same class. Moreover, the class of unlearning data is mainly influenced by the auxiliary dataset if $(1-\alpha)$ tends to 1. Therefore, Linear operation of Eq. \eqref{eq:lr-sample-unlearn} satisfies requirements \eqref{eq:rq1} and \eqref{eq:rq2}, which is also illustrated in Theorem \ref{thm:thm1}.

\subsubsection{Unlearning Class in FL}
Consider the scenario that only one client $k_0$ attempts to unlearn class $c$ data as $D^u_{k_0} = \{(x_{k_0,i}^u, c)\}_{i=1}^m$, there are the following three steps in FedAU:
 \begin{itemize}
    \item Client $k_0$ \textit{designs an the auxiliary dataset} $\calD^{'}_{k_0} = \calD^{u'}_{k_0} \cup \calD_{k_0}^{r'}$. Specifically, $\calD^{r'}_{k_0}$ is based on $\calD^r_{k_0}$ by modifying the label $y_{k_0,i}^r$ with label $c$ and $\calD^{u'}_{k_0} =\calD^{u}_{k_0}$ (see blue line 3-9 of Algo. \ref{algo:unlearn-sample});
  \item Then client $k_0$ \textit{learns the auxiliary unlearning module} $W^a_{k_0}$ according to the $\calD^{'}_{k_0}$ during the learning stage (see green line 10-15 of Algo. \ref{algo:unlearn-sample}).
 \item Finally, when the unlearning request is proposed, the unlearning model $\hat{W}$ can be obtained as:
    \begin{equation} \label{eq:lr-class-unlearn}
    \hat{W}  = W^l - \beta W^a_{k_0},
\end{equation}
where $\beta$ is a large coefficient (see red line 21 of Algo. \ref{algo:unlearn-sample}). 
\end{itemize}
In Fig. \ref{fig:method}(b), it can be observed that the subtraction operation does not affect the class of the remaining data. This is because the remaining dataset and the auxiliary dataset have different classes, resulting in the class of the subtrahend being preserved. Additionally, the unlearning data and auxiliary data share the same class (represented by the black point), allowing the class to be removed when the subtraction is performed. Therefore, the linear operation defined by Equation \eqref{eq:lr-class-unlearn} satisfies the requirements \eqref{eq:rq1} and \eqref{eq:rq2}, as also illustrated in Theorem \ref{thm:thm1}.  

\begin{rmk}
We provide the analysis on how the value of $\alpha$ influence the performance of the remaining data and unlearning data in Sect. \ref{subsec:exp-aba}.
\end{rmk}
\begin{algorithm}[H]
\caption{Unlearning Sample in FL (\colorbox{rgb:red!2,65;green!30,60;blue!20,125}{Learning Module}, \colorbox{rgb:red!2,65;green!30,90;blue!20,125}{Auxiliary Unlearning Module} and \colorbox{rgb:red!30,155;green!20,20;blue!20,30}{Linear Operation})}\label{algo:unlearn-sample}
\textbf{Input:} Communication rounds $T$, Client number $K$, dataset $\calD_{k_0}$ (remaining data $\calD_{k_0}^r$ and unlearning data $\calD_{k_0}^u$) for unlearning client $k_0$.
\begin{algorithmic}[1]
  \STATE Initialize the feature extractor $E$, unlearning learning module $W^l$ and auxiliary unlearning module $W_{k_0}^a$
\FOR{$t = 1, 2, \dots, T$} 
    \colorbox{rgb:red!2,65;green!30,60;blue!20,125}{
		\parbox{0.42\textwidth}{\vbox{\STATE \gray{$\triangleright$ \textit{Clients perform:}}
    \FOR{Client $k$ in $\{1,\dots,K\}$}
    \STATE Set $E_k = E$, $W_k^l = W^l$;
    \STATE Compute the learning loss $\tilde{\ell}=\ell(\calD_k; E_k, W_k^l)$;
    \STATE $W_{k}^l \longleftarrow W_k^l - \eta \nabla_{W_k^l}\tilde{\ell}$;
    \STATE $E_k \longleftarrow E_k - \eta \nabla_{E_k}\tilde{\ell}$;
    \ENDFOR }}}
     \colorbox{rgb:red!2,65;green!30,90;blue!20,125}{
		\parbox{0.42\textwidth}{\vbox{
  \STATE Let $W_{k_0}^a = W^a$;
  \STATE Set $\calD^u_{k_0} = (x_{k_0,i}^u, y^{u'}_{k_0,i} \sim U(1,C))$;
     \STATE Set $\calD^{r'}_{k_0} =\calD^{r}_{k_0}$;
    \STATE Set $\calD^{'}_{k_0} = \calD^{u'}_{k_0} \cup \calD_{k_0}^{r'}$;
    \STATE Compute the learning loss $\tilde{\ell}=\ell(\calD^{'}_{k_0} ; E_{k_0},W_{k_0}^a) $;
    \STATE $W_{k_0}^a \longleftarrow W_{k_0}^a - \eta \nabla{W_{k_0}^a}\tilde{\ell}$;
   }}} 
    \STATE Upload the $W_k^l$ and $E_k$ to the server;
\STATE \gray{$\triangleright$ \textit{The server performs:}}
\STATE The server aggregates $E$ and $W^l$ as: 
$    W^l = \frac{1}{K} (W_1^l+ \cdots + W_K^l); E = \frac{1}{K} (E_1+ \cdots + E_K)$;
\STATE The server distributes $E$ and $W^l$ to all clients.
\ENDFOR
\STATE \colorbox{rgb:red!30,155;green!20,20;blue!20,30}{\parbox{0.42\textwidth}{\vbox{The server implements unlearning process: 
\begin{equation*}
    \hat{W} = \alpha W^l + (1-\alpha) W_{k_0}^a 
\end{equation*}}}} \\
  \RETURN $E, \hat{W}$
\end{algorithmic}
\end{algorithm}
\hspace{0.4cm}
\vspace{-20pt}
\begin{algorithm}[ht]
\caption{Unlearning Class in FL (\colorbox{rgb:red!2,65;green!30,60;blue!20,125}{Learning Module}, \colorbox{rgb:red!2,65;green!30,90;blue!20,125}{Auxiliary Unlearning Module} and \colorbox{rgb:red!30,155;green!20,20;blue!20,30}{Linear Operation})}\label{algo:unlearn-class}
\textbf{Input:} Communication rounds $T$, $\calD_{k_0}$ including remaining data $\calD_{k_0}^r$ and unlearning data $\calD_{k_0}^u$ (the label of $\calD_{k_0}^u$ is $c$) for unlearning client $k_0$.
\begin{algorithmic}[1]
  \STATE Initialize the feature extractor $E$, unlearning learning module $W^l$ and auxiliary unlearning module $W_{k_0}^a$
\FOR{$t = 1, 2, \dots, T$} 
    \colorbox{rgb:red!2,65;green!30,60;blue!20,125}{
		\parbox{0.42\textwidth}{\vbox{\STATE \gray{$\triangleright$ \textit{Clients perform:}}
    \FOR{Client $k$ in $\{1,\dots,K\}$}
    \STATE Set $E_k = E$, $W_k^l = W^l$;
    \STATE Compute the learning loss $\tilde{\ell}=\ell(\calD_k; E_k, W_k^l)$;
    \STATE $W_{k}^l \longleftarrow W_k^l - \eta \nabla_{W_k^l}\tilde{\ell}$;
    \STATE $E_k \longleftarrow E_k - \eta \nabla_{E_k}\tilde{\ell}$;
    \ENDFOR }}}
     \colorbox{rgb:red!2,65;green!30,90;blue!20,125}{
		\parbox{0.42\textwidth}{\vbox{
    \STATE Let $W_{k_0}^a = W^a$;
\STATE Set $\calD^{u'}_{k_0} = \calD^u_{k_0}$;
\STATE Set $\calD^{r'}_{k_0} = (x_{k_0,i}^r, c ) $; 
\STATE Set $\calD^{'}_{k_0} = \calD^{u'}_{k_0} \cup \calD_{k_0}^{r'}$;
\STATE Compute the learning loss $\tilde{\ell}=\ell(\calD^{'}_{k_0} ; E_{k_0}, W_{k_0}^a,) $;
\STATE $W_{k_0}^a \longleftarrow W_{k_0}^a - \eta \nabla_{W_{k_0}^a}\tilde{\ell}$;
}}}
\STATE Upload the $W_k^l$ and $E_k$ to the server;

\STATE \gray{$\triangleright$ \textit{The server performs:}}
\STATE The server aggregates $E$ and $W^l$ as: 
$    W^l = \frac{1}{K} (W_1^l+ \cdots + W_K^l); E = \frac{1}{K} (E_1+ \cdots + E_K)$;
\STATE The server distributes $E$ and $W^l$ to all clients.
\ENDFOR

\STATE\colorbox{rgb:red!30,155;green!20,20;blue!20,30}{\parbox{0.42\textwidth}{\vbox{The server implements unlearning process: 
  \begin{equation*}
      \hat{W} =  W^l -\beta W^a_{k_0} 
  \end{equation*}}}} \\
  \RETURN $E, \hat{W}$
\end{algorithmic}
\end{algorithm}
\vspace{-6pt}

\subsubsection{Unlearning a Client in FL}
Unlearning a client represents an extreme case of unlearning samples, allowing us to leverage strategies used for unlearning samples. The key difference is that in the case of unlearning a client $k_0$, there is no remaining data from that client ($|\calD_{k_0}^r|=0$). As a result, the auxiliary unlearning module $W^a_{k_0}$ cannot learn from the data of other clients. To address this, we propose an improved updating strategy for the auxiliary unlearning module, which involves combining the knowledge learned from $\calD_{k_0}^{u'}$ and the original model $W^l$ for each epoch. Further details can be found in the Appendix B.

\subsubsection{Unlearning for Multiple Clients}
The proposed FedAU can also be applied into satisfying unlearning request for multiple clients without consuming extra time. Specifically, for unlearning class, each client in $\calC$ \textbf{privately learn} the $W^a_k, k\in \calC$ with the goal of optimizing Eq. \eqref{eq:aux-objective}.
Then all clients obtain the unlearning model $\hat{W}$ as:
\begin{equation*}
    \hat{W} =  W^l- \sum_{k\in \calC} \beta_k W^a_k.
\end{equation*}
The detailed algorithm and results shown in Appendix B.

For unlearning sample, multiple clients $\calC$ \textbf{collaboratively learn} the $W^a$ that aiming to optimize:
\begin{equation} \label{eq:aux-objective-co}
W^a = \argmin_W \sum_{k \in \calC} \sum_{(x_{k, i},y_{k, i}) \in \mathcal D_k^{'}}  \frac{\ell(F_{E,W}(x_{k, i}),y_{k, i})}{\sum_{k \in \calC} n_k}
\end{equation}
Then all clients obtain the unlearning model $\hat{W}$ as:
\begin{equation*}
    \hat{W} = \alpha W^l+ (1-\alpha)W^a
\end{equation*}
The detailed algorithm and results shown in Appendix B.

\begin{rmk}
In multiple client scenarios, the reason for the difference between unlearning samples and unlearning classes lies in the nature of the linear operations involved. When unlearning a class, the linear operation used is subtraction, which allows for the removal of multiple classes by subtracting $W^a_k$ for each client $k\in \calC$ individually. On the contrary, when unlearning samples, the operation is addition, where all $W^a_k$ for each client $k\in \calC$ are added together. This addition operation can potentially affect the unlearning effect because it is uncertain whether $W^a_{k_1}$ of client $k_1$ can effectively unlearn the unlearning samples $\calD_{k_2}^u$ of client $k_2$.

\end{rmk}



\subsection{Theoretical Analysis} \label{subsec:theo}
The following theorem demonstrates the proposed Algorithm \ref{algo:unlearn-sample} and \ref{algo:unlearn-class} satisfy Requirement \ref{eq:rq1} and \ref{eq:rq2} (see proof in Appendix D).  

\begin{theorem} \label{thm:thm1}
For client $k_0$ aims to remove $\calD^u_{k_0}$ from the $\calD_{k_0}$, and let $\calD^r_{k_0} = \calD_{k_0} - \calD^u_{k_0} $. There exist $\alpha$ and $\beta$ such that both unlearning Algorithm \ref{algo:unlearn-sample} and \ref{algo:unlearn-class} satisfy the requirement \eqref{eq:rq1} and \eqref{eq:rq2}, i.e., 
\begin{equation} \label{eq:upload-gradients}
\left\{
\begin{aligned}
&\argmax_iF^i_{W^l}(x) = \argmax_iF^i_{\hat{W}}(x),\quad x\in \calD^r_{k_0},\\
&\argmax_i F^i_{\hat{W}}(x) \neq y, \qquad \qquad \qquad (x,y)\in \calD^u_{k_0},\\
\end{aligned}
\right.
\end{equation}
\end{theorem}

Theorem \ref{thm:thm1} establishes the effectiveness of FedAU when a single client requests unlearning. Furthermore, we present a comprehensive theoretical analysis of the effectiveness of FedAU in scenarios where multiple clients request unlearning.


\begin{table*}[bp]
\centering
\renewcommand\arraystretch{1.6}
\scriptsize
\setlength{\tabcolsep}{0.7mm}
\begin{tabular}{@{}c|c||c|cc|cc|cc|cc|cc@{}}
\toprule
& & \multicolumn{1}{c|}{FedAvg} &  \multicolumn{2}{c|}{Retraining} & \multicolumn{2}{c|}{Amnesiac} & \multicolumn{2}{c|}{Class-disc} & \multicolumn{2}{c|}{FedEraser} & \multicolumn{2}{c}{FedAU} \\
\cline{3-13}
\multirow{-2}{*}{\begin{tabular}[c]{@{}c@{}}Dataset\\ (\%)\end{tabular}} & \multirow{-2}{*}{\begin{tabular}[c]{@{}c@{}}UL\\ Method\end{tabular}} & Rm-Acc & \multicolumn{2}{c|}{Ul/Rm-Acc} & \multicolumn{2}{c|}{Ul/Rm-Acc} & \multicolumn{2}{c|}{Ul/Rm-Acc} & \multicolumn{2}{c|}{Ul/Rm-Acc} & \multicolumn{2}{c}{Ul/Rm-Acc} \\
\midrule
& Samples &   87.77 $\pm$ 0.21  &  1.90 $\pm$ 0.20 & 87.36 $\pm$ 0.23 
&4.8 $\pm$ 0.99 &  85.91 $\pm$ 0.14 & --- & --- & --- & --- &  0.35 $\pm$ 0.07 &86.24$\pm$ 0.16 \\
& Classes & 87.50 $\pm$ 0.05 & 0.00 $\pm$ 0.00 & 87.45 $\pm$ 0.28 & 26.15 $\pm$ 2.76 & 74.93 $\pm$ 3.38 & 0.00 $\pm$ 0.00 & 79.42 $\pm$ 1.25 & --- & --- & 0.01 $\pm$ 0.01 & 87.71 $\pm$ 0.33 \\
\multirow{-3}{*}{\begin{tabular}[c]{@{}c@{}}CIFAR10\\ AlexNet\end{tabular}} & Clients & 87.49 $\pm$ 0.10 & 1.80 $\pm$ 0.16 & 87.33 $\pm$ 0.19 & 6.05 $\pm$ 0.35 & 75.34 $\pm$ 1.44 & --- & --- & 9.32 $\pm$ 0.11 & 84.60 $\pm$ 0.74 & 0.52 $\pm$ 0.06 & 86.83 $\pm$ 0.31 \\
\midrule
& Samples &  99.44 $\pm$ 0.02 & 0.44 $\pm$ 0.13 &  99.46 $\pm$ 0.04 
& 1.65 $\pm$ 0.07 &  98.87 $\pm$ 0.21 & --- & --- & --- & --- & 0.62 $\pm$ 0.23 & 99.36 $\pm$ 0.01  \\
& Classes & 99.50 $\pm$ 0.04 & 0.00 $\pm$ 0.00 & 99.54 $\pm$ 0.02 & 53.78 $\pm$ 6.06 & 57.95 $\pm$ 3.15 & 0.00 $\pm$ 0.00 & 99.13 $\pm$ 0.16 & --- & --- & 0.00 $\pm$ 0.00 & 99.63 $\pm$ 0.01 \\
\multirow{-3}{*}{\begin{tabular}[c]{@{}c@{}}MNIST\\ LeNet\end{tabular}} & Clients & 99.28 $\pm$ 0.06 & 0.77 $\pm$ 0.27 & 98.69 $\pm$ 0.04 & 0.53 $\pm$ 0.28 & 97.89 $\pm$ 0.47 & --- & --- & 8.05 $\pm$ 0.50 & 99.33 $\pm$ 0.21 & 0.66 $\pm$ 0.10 & 99.08 $\pm$ 0.12 \\
\bottomrule
\end{tabular}
\caption{The comparison with current methods, including FedAvg, Retraining, Amnesiac unlearning \protect\cite{graves2021amnesiac}, Class-dis \protect\cite{wang2022federated} and Federaser \protect\cite{liu2021federaser} and FedAU in different federated machine unlearning scenarios.}\label{tab:unlearn_result}
\end{table*}

\section{Experiment} \label{sec:exp}
\subsection{Experimental Setting}
\paragraph{\noindent\textbf{Models \& Datasets \& Setting.}}
We conduct experiments on three datasets:
\textit{MNIST} \cite{lecun2010mnist}, \textit{CIFAR10} and CIAFR100 \cite{krizhevsky2014cifar}. We adopt LeNet \cite{lecun1998gradient} for conducting experiments on MNIST and adopt \textit{AlexNet} \cite{NIPS2012_c399862d} on CIFAR10 and \textit{ResNet18} \cite{he2016deep} on CIFAR100. 

We simulate a HFL scenario consisting 10 clients under IID and Non-IID setting \cite{li2022federated} (following the Dirichlet distribution, $dir(\gamma)$). For unlearning samples, we employed the backdoor technique to generate the unlearning samples \cite{gao2022verifi}. The proportion of unlearning samples was set to 5\%, 10\%, and 20\% of the dataset. For unlearning a client, we considered scenarios where the data from the unlearning client accounted for 20\%, 50\%, and 100\% of the data from the other clients.
In addition, we conducted experiments involving unlearning for multiple clients. We varied the number of unlearning clients, exploring scenarios with 3, 5, 8, and 10 unlearning clients. Furthermore, we treated the last layer of the model as the auxiliary unlearning module. An ablation study on the position of the auxiliary unlearning module is provided in the Appendix B.
\paragraph{\noindent\textbf{The Baseline FMU Methods.}}
We compare six FMU methods, including Retraining/finetuning-based:  Retraining, FedEraser \cite{liu2021federaser}, Fedrecovery \cite{zhang2023fedrecovery}, gradient ascent-based: Amnesiac \cite{graves2021amnesiac},  Pruning-based: Class-dis \cite{wang2022federated} and the proposed FedAU  to evaluate the effectiveness.
\paragraph{\noindent\textbf{Evaluation Metrics.}}
We performed backdoor detection and membership inference attack (MIA) \cite{gao2022verifi,graves2021amnesiac} on the unlearned model to see if the influence of the targeted client was really removed by the proposed unlearning algorithm. The less the backdoor detection rate, and attack accuracy metric, the more effective the FMU methods are (Due to page limit, please refer the results of the recall in Appendix C). Moreover, the unlearning time cost and model performance of the remaining data is also utilized to evaluate all FMU methods. We refer all details about the experimental setting on Appendix A.

\subsection{Overall Evaluation}

\subsubsection{Evaluation of Unlearning Effect}
To ensure an effective unlearning method, it is crucial for the unlearned model to retain minimal information about the forgotten data. In Tab. \ref{tab:unlearn_result}, we present a comparison of accuracy of unlearning data achieved by various unlearning methods on the MNIST and CIFAR10 datasets. Our observations are as follows: 1) Amnesiac unlearning \cite{graves2021amnesiac} demonstrate strong performance in unlearning samples and clients, but exhibit lower effectiveness in unlearning classes (e.g., achieving a 20\% increase in the accuracy of unlearning data compared to the retraining method); 2) Class-dis \cite{wang2022federated} excel in unlearning classes, but are not suitable for unlearning samples and entire classes; 3) Our proposed method, FedAU, closely approximates the performance of retraining methods for unlearning samples, classes, and clients. For instance, the accuracy of unlearning data of FedAU is less than 1\% compared to retraining methods.

\subsubsection{Evaluation of Utility}
In Table \ref{tab:unlearn_result}, we evaluate the utility of the remaining data by measuring the remaining accuracy. The results indicate: 1) The Amnesiac unlearning method \cite{graves2021amnesiac} and the Federaser method \cite{liu2021federaser} are both affected in terms of remaining accuracy. For instance, the drop in remaining accuracy for the unlearning class in CIFAR10 using the Amnesiac unlearning method is more than 10\% compared to the retraining method;  2) Our proposed method, FedAU, effectively maintains the remaining accuracy with a minimal drop. Specifically, on CIFAR10, the drop in remaining accuracy is less than 1.5\%, and on MNIST, it is only 0.2\%.

\subsubsection{Evaluation of Time Cost}

\begin{table}[]
\centering
\small
\begin{tabular}{@{}cccc@{}}
\toprule
           & Samples  & Class    & Client   \\ \midrule
Retraining    & $\sim 10^3$    &  $\sim 10^3$    &  $\sim 10^3$   \\ \midrule
Amnesiac \cite{graves2021amnesiac}   & $\sim 10^0$      & $\sim 10^0$      & $\sim 10^0$   \\ \midrule
FedEraser \cite{liu2021federaser}  & /        & /        & $\sim 10^3$  \\ \midrule
Class-dis \cite{wang2022federated} & /        & $\sim 10^2$   & /        \\  \midrule
FedRecovery \cite{zhang2023fedrecovery} & /        & $\sim 10^0$   & /        \\  \midrule
FedAU (Ours)       & $\sim 10^{-3}$ & $\sim 10^{-3}$ & $\sim 10^{-3}$ \\ \bottomrule
\end{tabular}
\caption{Unlearning time cost (s) for different FMU methods under different federated machine unlearning scenarios.}\label{tab:time_cost}
\end{table}

Finally, we report the time consumed by each FMU (Fine-tuning Model Update) method to demonstrate the associated time costs (see details and more comparison on space consumption in Appendix C). The main results on the CIFAR10 dataset are presented in Tab. \ref{tab:time_cost}. From these results, we can draw two conclusions: 
\begin{enumerate}
    \item Among all the schemes, the Retraining scheme and schemes involving fine-tuning operations consume considerably more time compared to other methods;
    \item Although the Amnesiac and FedRecovery scheme requires a relatively small amount of time for unlearning, they still several orders of magnitude slower than FedAU;
    \item FedAU results in minimal additional training time,
e.g, additional 2s for AlexNet-CIFAR10.
This is because the AU module is lightweight such that training AU once consumes little time and training AU successfully only requires several epochs (see Appendix C). 
\end{enumerate}

\subsection{Ablation Study} \label{subsec:exp-aba}
This section introduces the ablation study on the some important factors: the number of unlearning clients, the Non-IID extent and the coefficient ($\alpha, \beta$) of FedAU. More ablation study on the proportion of unlearning samples, and impact of Coefficient $\alpha$ and $\beta$ see in Appendix C.
\subsubsection{Unlearning for Multiple Clients}
\begin{figure}[htbp]
    \centering
     \begin{subfigure}{0.22\textwidth}
  		 	  \includegraphics[width = \linewidth]{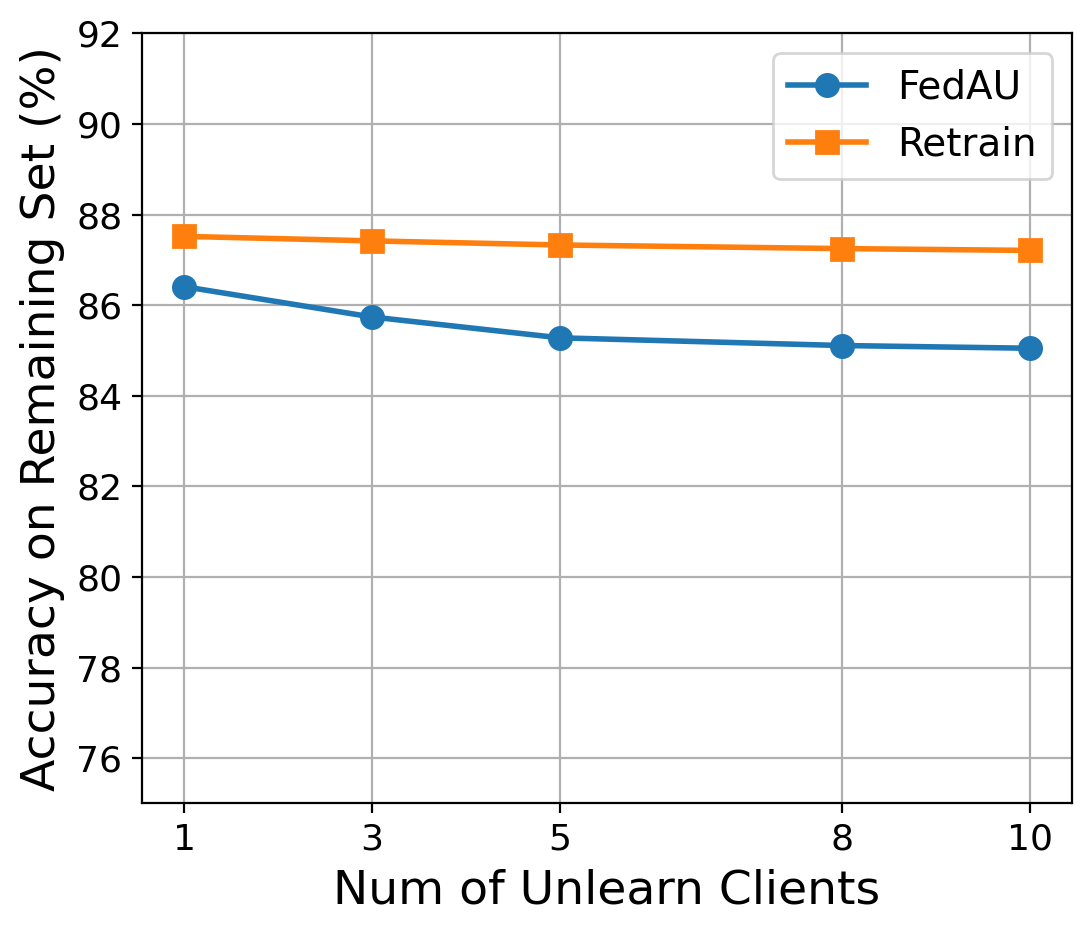}
    	\end{subfigure}
    \begin{subfigure}{0.22\textwidth}
  		 \includegraphics[width=1\textwidth]{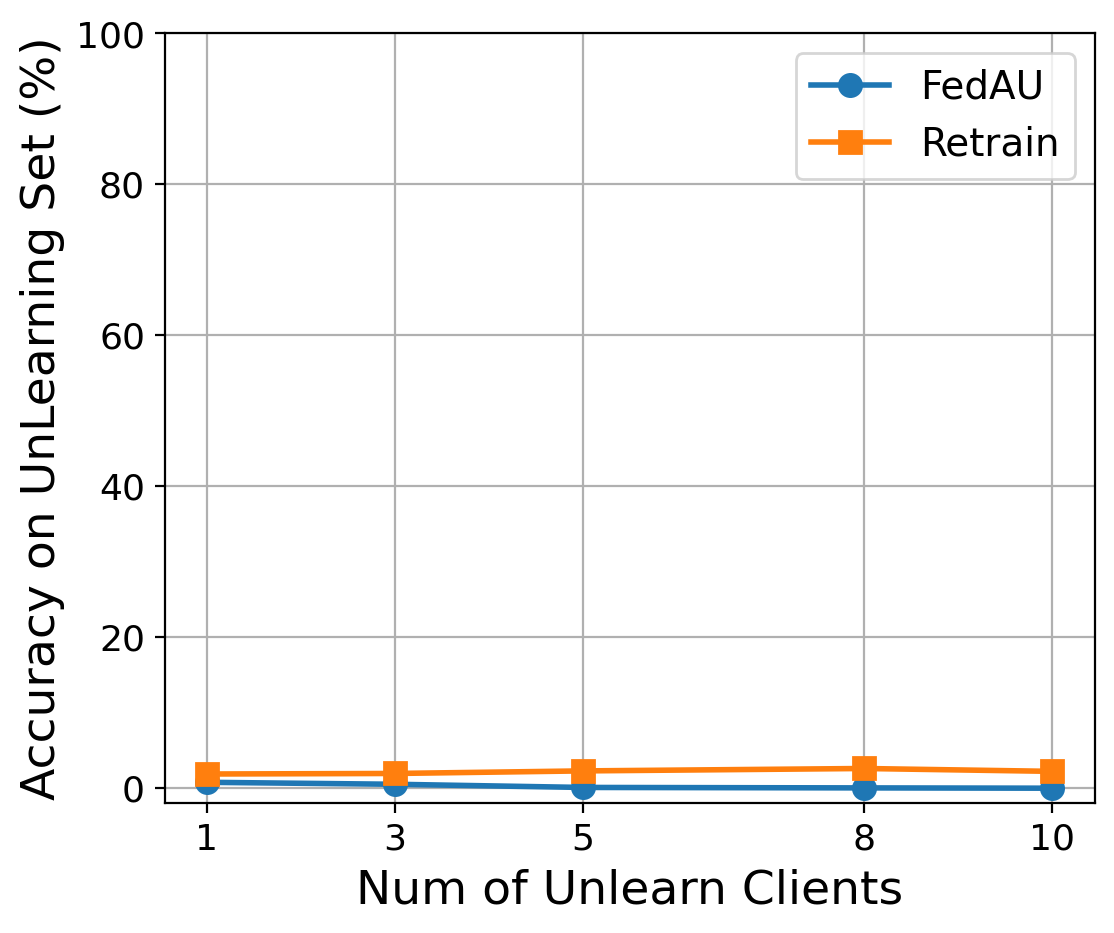}
    		\end{subfigure}
    \caption{The accuracy of FedAU and retraining methods on CIFAR10 with different number of unlearning clients.}
  
    \label{fig:mul-client}
\end{figure}

In the scenario where multiple clients request to unlearn, we allocate each client 
to unlearn 10\% of their respective datasets. The results in Figure \ref{fig:mul-client} depict the accuracy of unlearning and remaining data as the number of unlearning clients varies. The graph demonstrates that as the number of unlearning clients increases, the accuracy of the unlearning and remaining datasets achieved by our proposed method, FedAU, approaches that of the retraining method. This observation highlights the generality and effectiveness of our method in the multiple client unlearning scenario.

\subsubsection{Impact of Non-IID Extent}
In our study, we examined the impact of the Non-IID extent on the performance of the proposed FedAU (Federated Adaptive Unlearning) and retraining methods. To quantify the Non-IID extent, we used the $Dir(\gamma)$ distribution, where smaller values of $\gamma$ indicate more heterogeneous data.

\begin{figure}[htbp]
    \centering
     \begin{subfigure}{0.22\textwidth}
  		 	  \includegraphics[width = \linewidth]{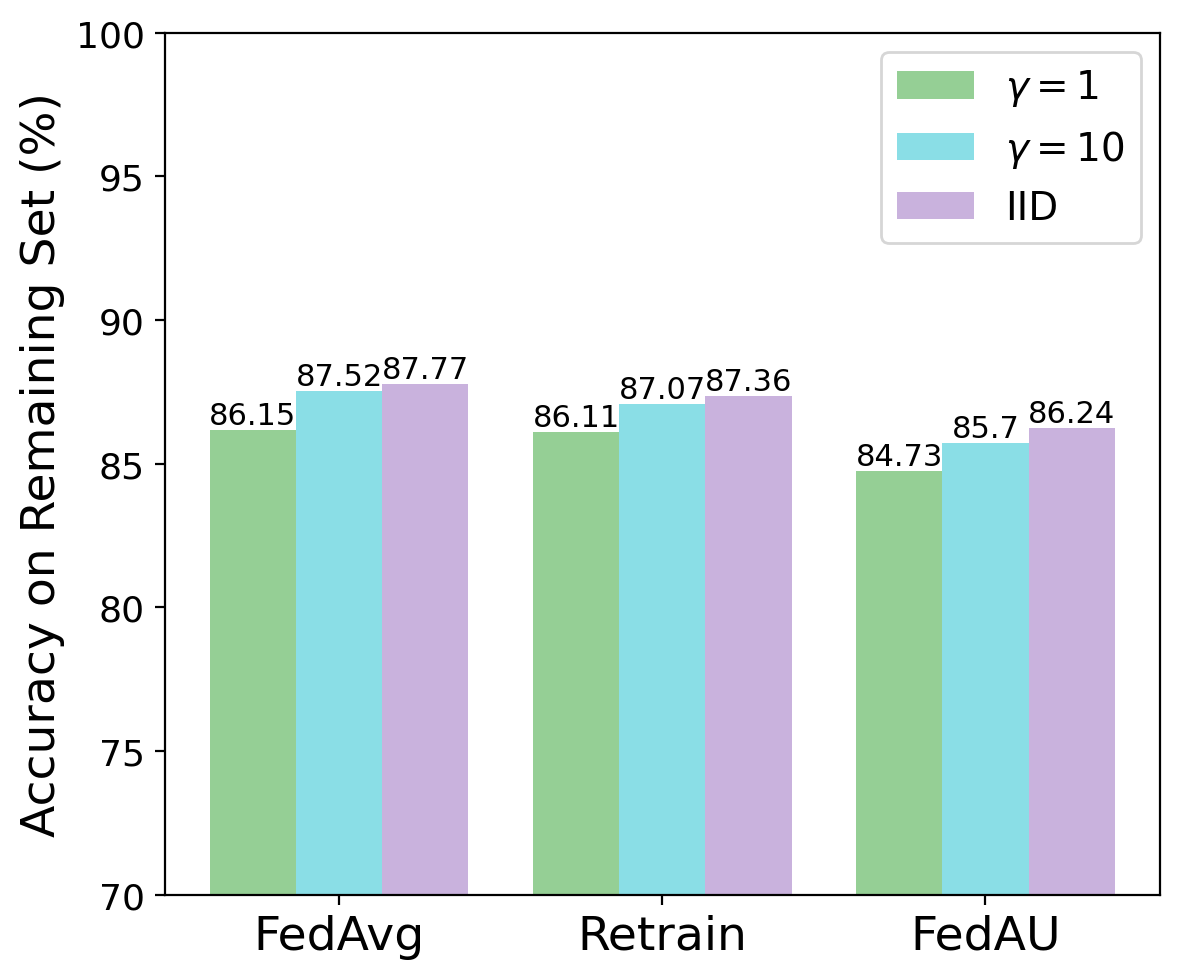}
    	\end{subfigure}
    \begin{subfigure}{0.22\textwidth}
  		 \includegraphics[width=1\textwidth]{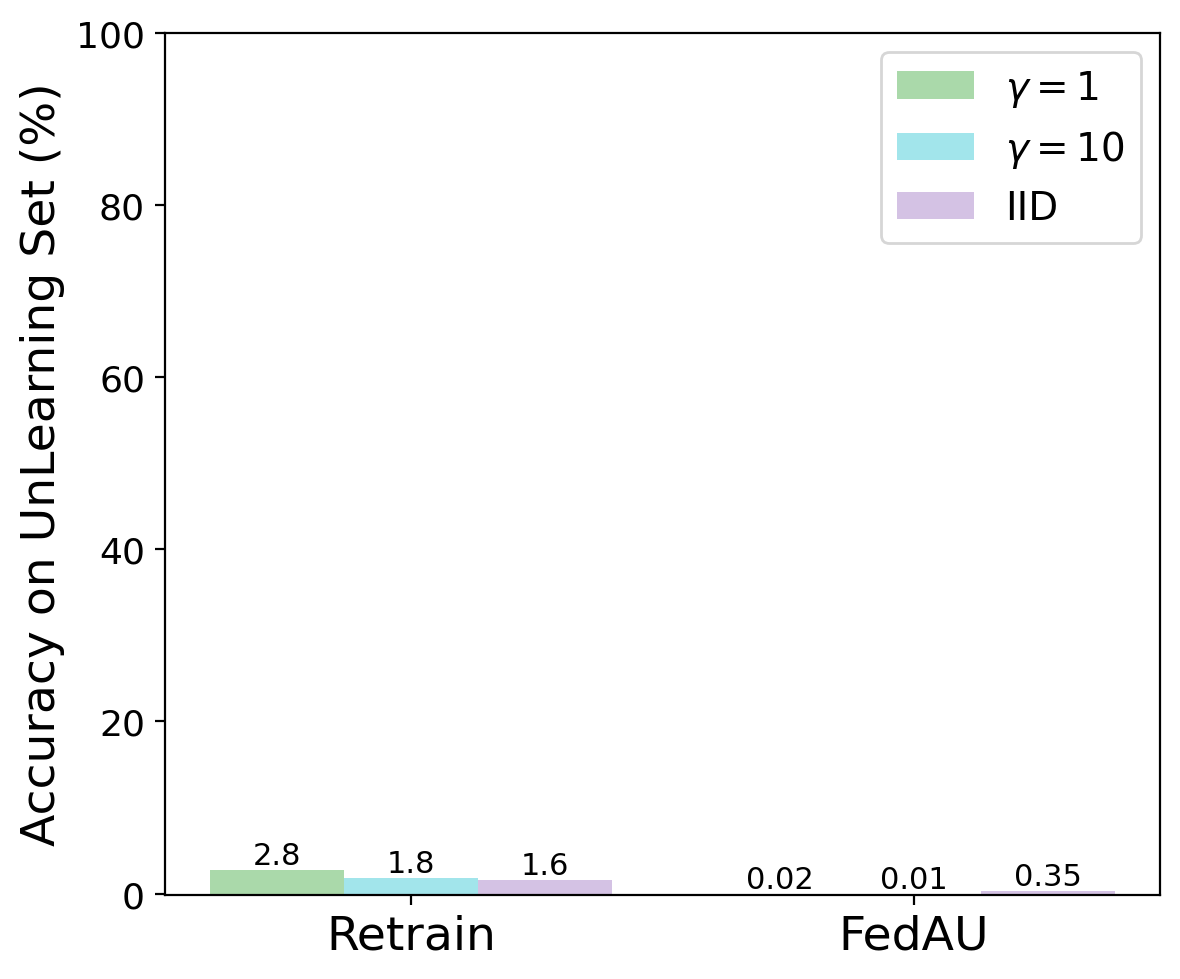}
    		\end{subfigure}
    \caption{The impact of Non-IID on CIFAR10 for the proposed FedAU and Retraining methods. }
  
    \label{fig:Non-IID}
    
\end{figure}

Fig. \ref{fig:Non-IID} illustrates the results of our experiments. We observed that the proposed FedAU method achieved a significant unlearning effect, as evidenced by the accuracy on the unlearning samples being less than 0.1\% when $\gamma=1$. Additionally, the FedAU method successfully maintained the model accuracy on the remaining data, with a drop of less than 2\% compared to the retraining method.


\section{Conclusion}

In response to the limitations associated with unlearning in Federated Learning (FL), we have introduced FedAU, an innovative and efficient Federated Machine Unlearning (FMU) framework. 
Briefly, FedAU integrates a lightweight auxiliary unlearning module into the learning process, employing a straightforward linear operation to streamline unlearning without the need for additional time-consuming steps. 
Moreover, FedAU empowers multiple clients to simultaneously perform unlearning tasks and supports unlearning at various levels of granularity, ranging from individual data samples to specific classes and even client-level unlearning.
We hope that its versatility and performance can make it a promising tool for future developments in the field.
\clearpage

\section*{Acknowledgements}
This work is partly supported by National Natural Science Foundation of China (NO.62206154), Shenzhen Startup Funding (No.QD2023014C), and supported by the National Research Foundation, Singapore and Infocomm Media Development Authority under its Trust Tech Funding Initiative (No. DTC-RGC-04).

\section*{Contribution Statement}
Hanlin Gu and Gongxi Zhu contributed equally to this work. Yuxing Han is 
the corresponding author.

\bibliographystyle{named}
\bibliography{unlearning, FL, attack, model}

\begin{thebibliography}{}

\bibitem[\protect\citeauthoryear{Abadi \bgroup \em et al.\egroup
  }{2016}]{abadi2016deep}
Martin Abadi, Andy Chu, Ian Goodfellow, H~Brendan McMahan, Ilya Mironov, Kunal
  Talwar, and Li~Zhang.
\newblock Deep learning with differential privacy.
\newblock In {\em Proceedings of the 2016 ACM SIGSAC conference on computer and
  communications security}, pages 308--318, 2016.

\bibitem[\protect\citeauthoryear{Aji and Heafield}{2017}]{aji2017sparse}
Alham~Fikri Aji and Kenneth Heafield.
\newblock Sparse communication for distributed gradient descent.
\newblock In {\em Proceedings of the 2017 Conference on Empirical Methods in
  Natural Language Processing}, pages 440--445, 2017.

\bibitem[\protect\citeauthoryear{Cheng \bgroup \em et al.\egroup
  }{2021}]{secureboost}
K.~Cheng, T.~Fan, Y.~Jin, Y.~Liu, T.~Chen, D.~Papadopoulos, and Q.~Yang.
\newblock Secureboost: A lossless federated learning framework.
\newblock {\em IEEE Intelligent Systems}, 36(06):87--98, nov 2021.

\bibitem[\protect\citeauthoryear{Dryden \bgroup \em et al.\egroup
  }{2016}]{dryden2016communication}
Nikoli Dryden, Tim Moon, Sam~Ade Jacobs, and Brian Van~Essen.
\newblock Communication quantization for data-parallel training of deep neural
  networks.
\newblock In {\em 2016 2nd Workshop on Machine Learning in HPC Environments
  (MLHPC)}, pages 1--8. IEEE, 2016.

\bibitem[\protect\citeauthoryear{Fan \bgroup \em et al.\egroup
  }{2019}]{fan2019rethinking}
Lixin Fan, Kam~Woh Ng, and Chee~Seng Chan.
\newblock Rethinking deep neural network ownership verification: Embedding
  passports to defeat ambiguity attacks.
\newblock {\em Advances in neural information processing systems}, 32, 2019.

\bibitem[\protect\citeauthoryear{Fan \bgroup \em et al.\egroup
  }{2020}]{fan2020rethinking}
Lixin Fan, Kam~Woh Ng, Ce~Ju, Tianyu Zhang, Chang Liu, Chee~Seng Chan, and
  Qiang Yang.
\newblock Rethinking privacy preserving deep learning: How to evaluate and
  thwart privacy attacks.
\newblock In {\em Federated Learning}, pages 32--50. Springer, 2020.

\bibitem[\protect\citeauthoryear{Fan \bgroup \em et al.\egroup
  }{2021}]{fan2021deepip}
Lixin Fan, Kam~Woh Ng, Chee~Seng Chan, and Qiang Yang.
\newblock Deepip: Deep neural network intellectual property protection with
  passports.
\newblock {\em IEEE Transactions on Pattern Analysis \& Machine Intelligence},
  (01):1--1, 2021.

\bibitem[\protect\citeauthoryear{Fu \bgroup \em et al.\egroup
  }{2022a}]{fu2022label}
Chong Fu, Xuhong Zhang, Shouling Ji, Jinyin Chen, Jingzheng Wu, Shanqing Guo,
  Jun Zhou, Alex~X Liu, and Ting Wang.
\newblock Label inference attacks against vertical federated learning.
\newblock In {\em 31st USENIX Security Symposium (USENIX Security 22), Boston,
  MA}, 2022.

\bibitem[\protect\citeauthoryear{Fu \bgroup \em et al.\egroup
  }{2022b}]{blindFL}
Fangcheng Fu, Huanran Xue, Yong Cheng, Yangyu Tao, and Bin Cui.
\newblock Blindfl: Vertical federated machine learning without peeking into
  your data.
\newblock In {\em Proceedings of the 2022 International Conference on
  Management of Data}, SIGMOD '22, page 1316–1330, New York, NY, USA, 2022.
  Association for Computing Machinery.

\bibitem[\protect\citeauthoryear{Gasc{\'o}n \bgroup \em et al.\egroup
  }{2016}]{gascon2016secure}
Adri{\`a} Gasc{\'o}n, Phillipp Schoppmann, Borja Balle, Mariana Raykova, Jack
  Doerner, Samee Zahur, and David Evans.
\newblock Secure linear regression on vertically partitioned datasets.
\newblock {\em IACR Cryptol. ePrint Arch.}, 2016:892, 2016.

\bibitem[\protect\citeauthoryear{Hardy \bgroup \em et al.\egroup
  }{2017}]{hardy2017private}
Stephen Hardy, Wilko Henecka, Hamish Ivey-Law, Richard Nock, Giorgio Patrini,
  Guillaume Smith, and Brian Thorne.
\newblock Private federated learning on vertically partitioned data via entity
  resolution and additively homomorphic encryption.
\newblock {\em arXiv preprint arXiv:1711.10677}, 2017.

\bibitem[\protect\citeauthoryear{He \bgroup \em et al.\egroup
  }{2016}]{he2016deep}
Kaiming He, Xiangyu Zhang, Shaoqing Ren, and Jian Sun.
\newblock Deep residual learning for image recognition.
\newblock In {\em Proceedings of the IEEE conference on computer vision and
  pattern recognition}, pages 770--778, 2016.

\bibitem[\protect\citeauthoryear{He \bgroup \em et al.\egroup
  }{2019}]{he2019model}
Zecheng He, Tianwei Zhang, and Ruby~B Lee.
\newblock Model inversion attacks against collaborative inference.
\newblock In {\em Proceedings of the 35th Annual Computer Security Applications
  Conference}, pages 148--162, 2019.

\bibitem[\protect\citeauthoryear{He \bgroup \em et al.\egroup
  }{2022}]{he2022hybrid}
Yuanqin He, Yan Kang, Jiahuan Luo, Lixin Fan, and Qiang Yang.
\newblock A hybrid self-supervised learning framework for vertical federated
  learning.
\newblock {\em arXiv preprint arXiv:2208.08934}, 2022.

\bibitem[\protect\citeauthoryear{Hu \bgroup \em et al.\egroup
  }{2022}]{hu2022vertical}
Yuzheng Hu, Tianle Cai, Jinyong Shan, Shange Tang, Chaochao Cai, Ethan Song,
  Bo~Li, and Dawn Song.
\newblock Is vertical logistic regression privacy-preserving? a comprehensive
  privacy analysis and beyond.
\newblock {\em arXiv preprint arXiv:2207.09087}, 2022.

\bibitem[\protect\citeauthoryear{Huang \bgroup \em et al.\egroup
  }{2020}]{huang2020instahide}
Yangsibo Huang, Zhao Song, Kai Li, and Sanjeev Arora.
\newblock Instahide: Instance-hiding schemes for private distributed learning.
\newblock In {\em International conference on machine learning}, pages
  4507--4518. PMLR, 2020.

\bibitem[\protect\citeauthoryear{Jin \bgroup \em et al.\egroup
  }{2021}]{jin2021cafe}
Xiao Jin, Pin-Yu Chen, Chia-Yi Hsu, Chia-Mu Yu, and Tianyi Chen.
\newblock Cafe: Catastrophic data leakage in vertical federated learning.
\newblock {\em NeurIPS}, 34:994--1006, 2021.

\bibitem[\protect\citeauthoryear{Kang \bgroup \em et al.\egroup
  }{2022a}]{kang2022prada}
Y.~Kang, Y.~He, J.~Luo, T.~Fan, Y.~Liu, and Q.~Yang.
\newblock Privacy-preserving federated adversarial domain adaptation over
  feature groups for interpretability.
\newblock {\em IEEE Transactions on Big Data}, (01):1--12, jul 2022.

\bibitem[\protect\citeauthoryear{Kang \bgroup \em et al.\egroup
  }{2022b}]{kangyan2022fedcvt}
Yan Kang, Yang Liu, and Xinle Liang.
\newblock {{FedCVT}}: {{Semi-supervised Vertical Federated Learning}} with
  {{Cross-view Training}}.
\newblock {\em ACM Transactions on Intelligent Systems and Technology (TIST)},
  May 2022.

\bibitem[\protect\citeauthoryear{Kang \bgroup \em et al.\egroup
  }{2022c}]{kang2022framework}
Yan Kang, Jiahuan Luo, Yuanqin He, Xiaojin Zhang, Lixin Fan, and Qiang Yang.
\newblock A framework for evaluating privacy-utility trade-off in vertical
  federated learning.
\newblock {\em arXiv preprint arXiv:2209.03885}, 2022.

\bibitem[\protect\citeauthoryear{Krizhevsky \bgroup \em et al.\egroup
  }{2012}]{NIPS2012_c399862d}
Alex Krizhevsky, Ilya Sutskever, and Geoffrey~E Hinton.
\newblock Imagenet classification with deep convolutional neural networks.
\newblock In F.~Pereira, C.J. Burges, L.~Bottou, and K.Q. Weinberger, editors,
  {\em Advances in Neural Information Processing Systems}, volume~25. Curran
  Associates, Inc., 2012.

\bibitem[\protect\citeauthoryear{Krizhevsky \bgroup \em et al.\egroup
  }{2014}]{krizhevsky2014cifar}
Alex Krizhevsky, Vinod Nair, and Geoffrey Hinton.
\newblock The cifar-10 dataset.
\newblock {\em online: http://www. cs. toronto. edu/kriz/cifar. html}, 55(5),
  2014.

\bibitem[\protect\citeauthoryear{LeCun \bgroup \em et al.\egroup
  }{1998}]{lecun1998gradient}
Yann LeCun, L{\'e}on Bottou, Yoshua Bengio, and Patrick Haffner.
\newblock Gradient-based learning applied to document recognition.
\newblock {\em Proceedings of the IEEE}, 86(11):2278--2324, 1998.

\bibitem[\protect\citeauthoryear{LeCun \bgroup \em et al.\egroup
  }{2010}]{lecun2010mnist}
Yann LeCun, Corinna Cortes, and CJ~Burges.
\newblock Mnist handwritten digit database.
\newblock {\em ATT Labs [Online]. Available: http://yann.lecun.com/exdb/mnist},
  2, 2010.

\bibitem[\protect\citeauthoryear{Li \bgroup \em et al.\egroup
  }{2020}]{li2020review}
Li~Li, Yuxi Fan, Mike Tse, and Kuo-Yi Lin.
\newblock A review of applications in federated learning.
\newblock {\em Computers \& Industrial Engineering}, 149:106854, 2020.

\bibitem[\protect\citeauthoryear{Li \bgroup \em et al.\egroup
  }{2022a}]{li2022fedipr}
Bowen Li, Lixin Fan, Hanlin Gu, Jie Li, and Qiang Yang.
\newblock Fedipr: Ownership verification for federated deep neural network
  models.
\newblock {\em IEEE Transactions on Pattern Analysis and Machine Intelligence},
  2022.

\bibitem[\protect\citeauthoryear{Li \bgroup \em et al.\egroup
  }{2022b}]{oscar2022split}
Oscar Li, Jiankai Sun, Xin Yang, Weihao Gao, Hongyi Zhang, Junyuan Xie,
  Virginia Smith, and Chong Wang.
\newblock Label leakage and protection in two-party split learning.
\newblock In {\em International Conference on Learning Representations}, 2022.

\bibitem[\protect\citeauthoryear{Lin \bgroup \em et al.\egroup
  }{2018}]{lin2018deep}
Yujun Lin, Song Han, Huizi Mao, Yu~Wang, and Bill Dally.
\newblock Deep gradient compression: Reducing the communication bandwidth for
  distributed training.
\newblock In {\em International Conference on Learning Representations}, 2018.

\bibitem[\protect\citeauthoryear{Liu \bgroup \em et al.\egroup
  }{2020}]{liu2020secure}
Yang Liu, Yan Kang, Chaoping Xing, Tianjian Chen, and Qiang Yang.
\newblock A secure federated transfer learning framework.
\newblock {\em IEEE Intelligent Systems}, 35(4):70--82, 2020.

\bibitem[\protect\citeauthoryear{Liu \bgroup \em et al.\egroup
  }{2021}]{liu2021defending}
Yang Liu, Zhihao Yi, Yan Kang, Yuanqin He, Wenhan Liu, Tianyuan Zou, and Qiang
  Yang.
\newblock Defending label inference and backdoor attacks in vertical federated
  learning.
\newblock {\em arXiv preprint arXiv:2112.05409}, 2021.

\bibitem[\protect\citeauthoryear{Liu \bgroup \em et al.\egroup
  }{2022a}]{liu2022vertical}
Yang Liu, Yan Kang, Tianyuan Zou, Yanhong Pu, Yuanqin He, Xiaozhou Ye,
  Ye~Ouyang, Ya-Qin Zhang, and Qiang Yang.
\newblock Vertical federated learning.
\newblock {\em arXiv preprint arXiv:2211.12814}, 2022.

\bibitem[\protect\citeauthoryear{Liu \bgroup \em et al.\egroup
  }{2022b}]{liu2022cross}
Yang Liu, Xinle Liang, Jiahuan Luo, Yuanqin He, Tianjian Chen, Quanming Yao,
  and Qiang Yang.
\newblock Cross-silo federated neural architecture search for heterogeneous and
  cooperative systems.
\newblock In {\em Federated and Transfer Learning}, pages 57--86. Springer,
  2022.

\bibitem[\protect\citeauthoryear{Luo \bgroup \em et al.\egroup
  }{2021}]{Luo2021fi}
Xinjian Luo, Yuncheng Wu, Xiaokui Xiao, and Beng~Chin Ooi.
\newblock Feature inference attack on model predictions in vertical federated
  learning.
\newblock In {\em 2021 {IEEE} 37th International Conference on Data Engineering
  ({ICDE})}. {IEEE}, apr 2021.

\bibitem[\protect\citeauthoryear{Wu \bgroup \em et al.\egroup
  }{2015}]{wu20153d}
Zhirong Wu, Shuran Song, Aditya Khosla, Fisher Yu, Linguang Zhang, Xiaoou Tang,
  and Jianxiong Xiao.
\newblock 3d shapenets: A deep representation for volumetric shapes.
\newblock In {\em Proceedings of the IEEE conference on computer vision and
  pattern recognition}, pages 1912--1920, 2015.

\bibitem[\protect\citeauthoryear{Yang \bgroup \em et al.\egroup
  }{2019}]{yang2019federated}
Qiang Yang, Yang Liu, Tianjian Chen, and Yongxin Tong.
\newblock Federated machine learning: Concept and applications.
\newblock {\em ACM Transactions on Intelligent Systems and Technology (TIST)},
  10(2):1--19, 2019.

\bibitem[\protect\citeauthoryear{Zhang \bgroup \em et al.\egroup
  }{2018}]{zhang2018mixup}
Hongyi Zhang, Moustapha Cisse, Yann~N Dauphin, and David Lopez-Paz.
\newblock mixup: Beyond empirical risk minimization.
\newblock In {\em International Conference on Learning Representations}, 2018.

\bibitem[\protect\citeauthoryear{Zhang \bgroup \em et al.\egroup
  }{2022}]{zhang2022trading}
Xiaojin Zhang, Yan Kang, Kai Chen, Lixin Fan, and Qiang Yang.
\newblock Trading off privacy, utility and efficiency in federated learning.
\newblock {\em arXiv preprint arXiv:2209.00230}, 2022.

\bibitem[\protect\citeauthoryear{Zhu \bgroup \em et al.\egroup
  }{2019}]{zhu2019dlg}
Ligeng Zhu, Zhijian Liu, , and Song Han.
\newblock Deep leakage from gradients.
\newblock In {\em Annual Conference on Neural Information Processing Systems
  (NeurIPS)}, 2019.

\bibitem[\protect\citeauthoryear{Zou \bgroup \em et al.\egroup
  }{2022}]{zou2022defending}
Tianyuan Zou, Yang Liu, Yan Kang, Wenhan Liu, Yuanqin He, Zhihao Yi, Qiang
  Yang, and Ya-Qin Zhang.
\newblock Defending batch-level label inference and replacement attacks in
  vertical federated learning.
\newblock {\em IEEE Transactions on Big Data}, 2022.

\end{thebibliography}


\begin{thebibliography}{}

\bibitem[\protect\citeauthoryear{Bourtoule \bgroup \em et al.\egroup }{2021}]{bourtoule2021machine}
Lucas Bourtoule, Varun Chandrasekaran, Christopher~A Choquette-Choo, Hengrui Jia, Adelin Travers, Baiwu Zhang, David Lie, and Nicolas Papernot.
\newblock Machine unlearning.
\newblock In {\em 2021 IEEE Symposium on Security and Privacy (SP)}, pages 141--159. IEEE, 2021.

\bibitem[\protect\citeauthoryear{Che \bgroup \em et al.\egroup }{2023}]{che2023fast}
Tianshi Che, Yang Zhou, Zijie Zhang, Lingjuan Lyu, Ji~Liu, Da~Yan, Dejing Dou, and Jun Huan.
\newblock Fast federated machine unlearning with nonlinear functional theory.
\newblock In {\em International conference on machine learning}, pages 4241--4268. PMLR, 2023.

\bibitem[\protect\citeauthoryear{Chen \bgroup \em et al.\egroup }{2023}]{chen2023boundary}
Min Chen, Weizhuo Gao, Gaoyang Liu, Kai Peng, and Chen Wang.
\newblock Boundary unlearning: Rapid forgetting of deep networks via shifting the decision boundary.
\newblock In {\em Proceedings of the IEEE/CVF Conference on Computer Vision and Pattern Recognition}, pages 7766--7775, 2023.

\bibitem[\protect\citeauthoryear{Cheng \bgroup \em et al.\egroup }{2020}]{cheng2020federated}
Yong Cheng, Yang Liu, Tianjian Chen, and Qiang Yang.
\newblock Federated learning for privacy-preserving ai.
\newblock {\em Communications of the ACM}, 63(12):33--36, 2020.

\bibitem[\protect\citeauthoryear{Deng \bgroup \em et al.\egroup }{2020}]{deng2020distributionally}
Yuyang Deng, Mohammad~Mahdi Kamani, and Mehrdad Mahdavi.
\newblock Distributionally robust federated averaging.
\newblock {\em Advances in neural information processing systems}, 33:15111--15122, 2020.

\bibitem[\protect\citeauthoryear{Dwork}{2006}]{dwork2006differential}
Cynthia Dwork.
\newblock Differential privacy.
\newblock In {\em International colloquium on automata, languages, and programming}, pages 1--12. Springer, 2006.

\bibitem[\protect\citeauthoryear{Eisenhofer \bgroup \em et al.\egroup }{2022}]{Eisenhofer2022Verifiable}
Thorsten Eisenhofer, Doreen Riepel, Varun Chandrasekaran, Esha Ghosh, O.~Ohrimenko, and Nicolas Papernot.
\newblock Verifiable and provably secure machine unlearning.
\newblock {\em ArXiv}, abs/2210.09126, 2022.

\bibitem[\protect\citeauthoryear{Gao \bgroup \em et al.\egroup }{2022}]{gao2022verifi}
Xiangshan Gao, Xingjun Ma, Jingyi Wang, Youcheng Sun, Bo~Li, Shouling Ji, Peng Cheng, and Jiming Chen.
\newblock Verifi: Towards verifiable federated unlearning.
\newblock {\em arXiv preprint arXiv:2205.12709}, 2022.

\bibitem[\protect\citeauthoryear{Goldreich}{1998}]{goldreich1998secure}
Oded Goldreich.
\newblock Secure multi-party computation.
\newblock {\em Manuscript. Preliminary version}, 78(110):1--108, 1998.

\bibitem[\protect\citeauthoryear{Graves \bgroup \em et al.\egroup }{2021}]{graves2021amnesiac}
Laura Graves, Vineel Nagisetty, and Vijay Ganesh.
\newblock Amnesiac machine learning.
\newblock In {\em Proceedings of the AAAI Conference on Artificial Intelligence}, volume~35, pages 11516--11524, 2021.

\bibitem[\protect\citeauthoryear{Halimi \bgroup \em et al.\egroup }{2022}]{halimi2022federated}
Anisa Halimi, Swanand~Ravindra Kadhe, Ambrish Rawat, and Nathalie~Baracaldo Angel.
\newblock Federated unlearning: How to efficiently erase a client in fl?
\newblock In {\em International Conference on Machine Learning}, 2022.

\bibitem[\protect\citeauthoryear{Harding \bgroup \em et al.\egroup }{2019}]{harding2019understanding}
Elizabeth~Liz Harding, Jarno~J Vanto, Reece Clark, L~Hannah~Ji, and Sara~C Ainsworth.
\newblock Understanding the scope and impact of the california consumer privacy act of 2018.
\newblock {\em Journal of Data Protection \& Privacy}, 2(3):234--253, 2019.

\bibitem[\protect\citeauthoryear{He \bgroup \em et al.\egroup }{2016}]{he2016deep}
Kaiming He, Xiangyu Zhang, Shaoqing Ren, and Jian Sun.
\newblock Deep residual learning for image recognition.
\newblock In {\em Proceedings of the IEEE conference on computer vision and pattern recognition}, pages 770--778, 2016.

\bibitem[\protect\citeauthoryear{He \bgroup \em et al.\egroup }{2019}]{he2019model}
Zecheng He, Tianwei Zhang, and Ruby~B Lee.
\newblock Model inversion attacks against collaborative inference.
\newblock In {\em Proceedings of the 35th Annual Computer Security Applications Conference}, pages 148--162, 2019.

\bibitem[\protect\citeauthoryear{Jose and Simeone}{2021}]{Jose2021A}
Sharu~Theresa Jose and O.~Simeone.
\newblock A unified pac-bayesian framework for machine unlearning via information risk minimization.
\newblock {\em 2021 IEEE 31st International Workshop on Machine Learning for Signal Processing (MLSP)}, pages 1--6, 2021.

\bibitem[\protect\citeauthoryear{Kone{\v{c}}n{\`y} \bgroup \em et al.\egroup }{2015}]{konevcny2015federated}
Jakub Kone{\v{c}}n{\`y}, Brendan McMahan, and Daniel Ramage.
\newblock Federated optimization: Distributed optimization beyond the datacenter.
\newblock {\em arXiv preprint arXiv:1511.03575}, 2015.

\bibitem[\protect\citeauthoryear{Krizhevsky \bgroup \em et al.\egroup }{2012}]{NIPS2012_c399862d}
Alex Krizhevsky, Ilya Sutskever, and Geoffrey~E Hinton.
\newblock Imagenet classification with deep convolutional neural networks.
\newblock In F.~Pereira, C.J. Burges, L.~Bottou, and K.Q. Weinberger, editors, {\em Advances in Neural Information Processing Systems}, volume~25. Curran Associates, Inc., 2012.

\bibitem[\protect\citeauthoryear{Krizhevsky \bgroup \em et al.\egroup }{2014}]{krizhevsky2014cifar}
Alex Krizhevsky, Vinod Nair, and Geoffrey Hinton.
\newblock The cifar-10 dataset.
\newblock {\em online: http://www. cs. toronto. edu/kriz/cifar. html}, 55(5), 2014.

\bibitem[\protect\citeauthoryear{LeCun \bgroup \em et al.\egroup }{1998}]{lecun1998gradient}
Yann LeCun, L{\'e}on Bottou, Yoshua Bengio, and Patrick Haffner.
\newblock Gradient-based learning applied to document recognition.
\newblock {\em Proceedings of the IEEE}, 86(11):2278--2324, 1998.

\bibitem[\protect\citeauthoryear{LeCun \bgroup \em et al.\egroup }{2010}]{lecun2010mnist}
Yann LeCun, Corinna Cortes, and CJ~Burges.
\newblock Mnist handwritten digit database.
\newblock {\em ATT Labs [Online]. Available: http://yann.lecun.com/exdb/mnist}, 2, 2010.

\bibitem[\protect\citeauthoryear{Li \bgroup \em et al.\egroup }{2022}]{li2022federated}
Qinbin Li, Yiqun Diao, Quan Chen, and Bingsheng He.
\newblock Federated learning on non-iid data silos: An experimental study.
\newblock In {\em 2022 IEEE 38th International Conference on Data Engineering (ICDE)}, pages 965--978. IEEE, 2022.

\bibitem[\protect\citeauthoryear{Liu \bgroup \em et al.\egroup }{2021}]{liu2021federaser}
Gaoyang Liu, Xiaoqiang Ma, Yang Yang, Chen Wang, and Jiangchuan Liu.
\newblock Federaser: Enabling efficient client-level data removal from federated learning models.
\newblock In {\em 2021 IEEE/ACM 29th International Symposium on Quality of Service (IWQOS)}, pages 1--10. IEEE, 2021.

\bibitem[\protect\citeauthoryear{Liu \bgroup \em et al.\egroup }{2022a}]{liu2022right}
Yi~Liu, Lei Xu, Xingliang Yuan, Cong Wang, and Bo~Li.
\newblock The right to be forgotten in federated learning: An efficient realization with rapid retraining.
\newblock In {\em IEEE INFOCOM 2022-IEEE Conference on Computer Communications}, pages 1749--1758. IEEE, 2022.

\bibitem[\protect\citeauthoryear{Liu \bgroup \em et al.\egroup }{2022b}]{Liu2022The}
Yi~Liu, Lei Xu, Xingliang Yuan, Cong Wang, and Bo~Li.
\newblock The right to be forgotten in federated learning: An efficient realization with rapid retraining.
\newblock {\em IEEE INFOCOM 2022 - IEEE Conference on Computer Communications}, pages 1749--1758, 2022.

\bibitem[\protect\citeauthoryear{Liu \bgroup \em et al.\egroup }{2023}]{liu2023survey}
Ziyao Liu, Yu~Jiang, Jiyuan Shen, Minyi Peng, Kwok-Yan Lam, and Xingliang Yuan.
\newblock A survey on federated unlearning: Challenges, methods, and future directions.
\newblock {\em arXiv preprint arXiv:2310.20448}, 2023.

\bibitem[\protect\citeauthoryear{McMahan \bgroup \em et al.\egroup }{2017}]{mcmahan2017communication}
Brendan McMahan, Eider Moore, Daniel Ramage, Seth Hampson, and Blaise~Aguera y~Arcas.
\newblock Communication-efficient learning of deep networks from decentralized data.
\newblock In {\em Artificial Intelligence and Statistics}, pages 1273--1282. PMLR, 2017.

\bibitem[\protect\citeauthoryear{Mercuri \bgroup \em et al.\egroup }{2022}]{mercuri2022introduction}
Salvatore Mercuri, Raad Khraishi, Ramin Okhrati, Devesh Batra, Conor Hamill, Taha Ghasempour, and Andrew Nowlan.
\newblock An introduction to machine unlearning.
\newblock {\em arXiv preprint arXiv:2209.00939}, 2022.

\bibitem[\protect\citeauthoryear{Nasr \bgroup \em et al.\egroup }{2019}]{nasr2019comprehensive}
Milad Nasr, Reza Shokri, and Amir Houmansadr.
\newblock Comprehensive privacy analysis of deep learning: Passive and active white-box inference attacks against centralized and federated learning.
\newblock In {\em 2019 IEEE symposium on security and privacy (SP)}, pages 739--753. IEEE, 2019.

\bibitem[\protect\citeauthoryear{Sekhari \bgroup \em et al.\egroup }{2021}]{sekhari2021remember}
Ayush Sekhari, Jayadev Acharya, Gautam Kamath, and Ananda~Theertha Suresh.
\newblock Remember what you want to forget: Algorithms for machine unlearning.
\newblock {\em Advances in Neural Information Processing Systems}, 34:18075--18086, 2021.

\bibitem[\protect\citeauthoryear{Su and Li}{2023}]{su2023asynchronous}
Ningxin Su and Baochun Li.
\newblock Asynchronous federated unlearning.
\newblock In {\em IEEE INFOCOM 2023-IEEE Conference on Computer Communications}, pages 1--10. IEEE, 2023.

\bibitem[\protect\citeauthoryear{Sun \bgroup \em et al.\egroup }{2022}]{sun2022decentralized}
Tao Sun, Dongsheng Li, and Bao Wang.
\newblock Decentralized federated averaging.
\newblock {\em IEEE Transactions on Pattern Analysis and Machine Intelligence}, 45(4):4289--4301, 2022.

\bibitem[\protect\citeauthoryear{Wang \bgroup \em et al.\egroup }{2022}]{wang2022federated}
Junxiao Wang, Song Guo, Xin Xie, and Heng Qi.
\newblock Federated unlearning via class-discriminative pruning.
\newblock In {\em Proceedings of the ACM Web Conference 2022}, pages 622--632, 2022.

\bibitem[\protect\citeauthoryear{Wu \bgroup \em et al.\egroup }{2022}]{wu2022federated}
Chen Wu, Sencun Zhu, and Prasenjit Mitra.
\newblock Federated unlearning with knowledge distillation.
\newblock {\em arXiv preprint arXiv:2201.09441}, 2022.

\bibitem[\protect\citeauthoryear{Yang \bgroup \em et al.\egroup }{2019}]{yang2019federated}
Qiang Yang, Yang Liu, Tianjian Chen, and Yongxin Tong.
\newblock Federated machine learning: Concept and applications.
\newblock {\em ACM Transactions on Intelligent Systems and Technology (TIST)}, 10(2):1--19, 2019.

\bibitem[\protect\citeauthoryear{Zhang \bgroup \em et al.\egroup }{2023}]{zhang2023fedrecovery}
Lefeng Zhang, Tianqing Zhu, Haibin Zhang, Ping Xiong, and Wanlei Zhou.
\newblock Fedrecovery: Differentially private machine unlearning for federated learning frameworks.
\newblock {\em IEEE Transactions on Information Forensics and Security}, 2023.

\end{thebibliography}

\newpage
\appendix

\noindent \textbf{\huge Appendix}
\vspace{4pt}

We provide the experimental setting, detailed algorithms, additional experiments and proof in this appendix. 
\section{Experimental Setting}
\begin{table*}[bp]  \renewcommand\arraystretch{0.8}
	\resizebox{1\textwidth}{!}{
		\begin{tabular}{l|ccc}
			\hline
			Hyper-parameter & LeNet-MNIST& AlexNet-CIFAR10 & ResNet-CIFAR100 \\ \hline
			Optimization method & SGD & SGD& SGD\\
			Learning rate & 1e-2 & 1e-2& 1e-2\\
			Weight decay & 4e-5 & 4e-5& 4e-5\\
			Batch size & 32 & 32& 128\\
			Iterations & 100 & 200& 200 \\
   The proportion of unlearning samples &10\%&[5\%, 20\%]&10\% \\
   The number of unlearning clients &1&[1-10]&1 \\
   Coefficient $\alpha$&0.9&[0.6-0.99]&0.9 \\
   Coefficient $\beta$ &1&[0.1-3.0]&1 \\
   Position of $W^a_{k_0}$ &last three layers&last layer&last layer\\
			\hline
	\end{tabular}}
	
	\caption{Hyper-parameters used for training in FedAU. }
	\label{tab:train-params-app}
\end{table*}
\paragraph{\noindent\textbf{Models \& Dataset.}}
We conduct experiments on three datasets:
\textit{MNIST} \cite{lecun2010mnist}, \textit{CIFAR10} and CIFAR100 \cite{krizhevsky2014cifar}. We adopt LeNet \cite{lecun1998gradient} for conducting experiments on MNIST and adopt \textit{AlexNet} \cite{NIPS2012_c399862d} on CIFAR10 and \textit{ResNet18} \cite{he2016deep} on CIFAR100. Furthermore, we treated the last layer of the model as the auxiliary unlearning module. An ablation study on the position of the auxiliary unlearning module is provided in the Sect. \ref{subsec:app6}. See more details of parameters in Tab. \ref{tab:train-params-app}.

\paragraph{\noindent\textbf{FL Setting.}} 
We simulate a HFL scenario consisting 10 clients under IID and Non-IID setting \cite{li2022federated}, where the clients' data distribution follows the Dirichlet distribution ($dir(\gamma)$). The small $\gamma$ indicates the large heterogeneity. And We set $\gamma$ to 1, 10 and $\infty$ (IID). 

\paragraph{\noindent\textbf{Unlearning Setting.}} 
For unlearning samples, we employed the backdoor technique to generate the unlearning samples \cite{gao2022verifi}. The proportion of unlearning samples was set to 5\%, 10\%, and 20\% of the dataset. For unlearning a client, we considered scenarios where the data from the unlearning client accounted for 20\%, 50\%, and 100\% of the data from the other clients.
In addition, we conducted experiments involving unlearning for multiple clients, where multiple clients request to unlearn 10\% samples. We varied the number of unlearning clients, exploring scenarios with 3, 5, 8, and 10 unlearning clients. 

\paragraph{\noindent\textbf{The Baseline FMU Methods.}}
We compare six FMU methods, including Retraining/finetuning-based:  Retraining, FedEraser \cite{liu2021federaser}, FedRecovery \cite{zhang2023fedrecovery}, gradient ascent-based: Amnesiac \cite{graves2021amnesiac},  Pruning-based: Class-dis \cite{wang2022federated} and the proposed FedAU  to evaluate the effectiveness.

\paragraph{\noindent\textbf{Evaluation Metrics.}}
We consider the three objectives to evaluate the difference FMU methods in this paper: \textbf{Accuracy of Remaining Data}, \textbf{Unlearning Effect} and \textbf{Cost}. 
\begin{itemize}
    \item Firstly, we use the test accuracy on the remaining data (Rm-Acc); 
    \item Secondly, we leverage two metrics to evaluate the unlearning effect. One is testing the accuracy on the unlearning data (Ul-Acc) \cite{gao2022verifi}. The smaller the Ul-Acc is, the better the unlearning effect is.  The other metric is utilizing the membership inference attack (MIA) by determining whether the unlearning data is the training data of the unlearning model. Moreover, we use the attack accuracy \cite{graves2021amnesiac} and the low attack accuracy indicates the good unlearning effect;
    \item Thirdly, we compute the unlearning time cost and the memory cost on the unlearning step for different FMU methods.

\end{itemize}

\section{Unlearning for Multiple Clients}
The proposed FedAU can also be applied into satisfying unlearning request for multiple clients without consuming extra time. This section illustrates the algorithm on how to unlearn sample and class for multiple clients (see Algo. \ref{algo:unlearn-sample-mul} and \ref{algo:unlearn-class-mul})
\subsection{Unlearning Class for Multiple Clients}
For unlearning class, each client in $\calC$ \textbf{privately learn} the $W^a_{k_0}, k_0\in \calC$ with the goal of optimizing Eq. \eqref{eq:aux-objective}, which is shown in line 10-17 of Algo. \ref{algo:unlearn-class-mul}. Then In the unlearning step, unlearning model $\hat{W}$ is obtained by the server as (see line 23 of Algo. \ref{algo:unlearn-class-mul}):
\begin{equation*}
    \hat{W} =  W^l- \sum_{k_0\in \calC} \beta_k W^a_{k_0}.
\end{equation*}

\subsection{Unlearning Sample for Multiple Clients}
For unlearning sample, multiple clients $\calC$ \textbf{collaboratively learn} the $W^a$ that aiming to optimize:
\begin{equation} \label{eq:aux-objective-co-app}
W^a = \argmin_W \sum_{k \in \calC} \sum_{(x_{k, i},y_{k, i}) \in \mathcal D_k^{'}}  \frac{\ell(F_{E,W}(x_{k, i}),y_{k, i})}{\sum_{k \in \calC} n_k}.
\end{equation}
Specifically, in each communication rounds, all unlearning clients learn their own auxiliary unlearning module $W_{k_0}^a$ as shown in line 10-17 of Algo. \ref{algo:unlearn-sample-mul}. And they upload all $W_{k_0}^a$ to the server to aggregate as shown in line 18-20 of Algo. \ref{algo:unlearn-class-mul}. Finally, unlearning model $\hat{W}$ is obtained by the server as (see line 23 of Algo. \ref{algo:unlearn-sample-mul}):
\begin{equation*}
    \hat{W} = \alpha W^l+ (1-\alpha)W^a
\end{equation*}

\begin{rmk}
In multiple client scenarios, the reason for the difference between unlearning samples and unlearning classes lies in the nature of the linear operations involved. When unlearning a class, the linear operation used is subtraction, which allows for the removal of multiple classes by subtracting $W^a_k$ for each client $k\in \calC$ individually. On the contrary, when unlearning samples, the operation is addition, where all $W^a_k$ for each client $k\in \calC$ are added together. This addition operation can potentially affect the unlearning effect because it is uncertain whether $W^a_{k_1}$ of client $k_1$ can effectively unlearn the unlearning samples $\calD_{k_2}^u$ of client $k_2$.

\end{rmk}

\begin{figure*}[t]
\vspace{-0.5cm}
\begin{minipage}{0.48\textwidth}
\begin{algorithm}[H]
\small
\caption{Unlearning Sample in FL for multiple clients (\colorbox{rgb:red!2,65;green!30,60;blue!20,125}{Learning Module}, \colorbox{rgb:red!2,65;green!30,90;blue!20,125}{Auxiliary Unlearning Module} and \colorbox{rgb:red!30,155;green!20,20;blue!20,30}{Linear Operation})}\label{algo:unlearn-sample-mul}
\textbf{Input:} Communication rounds $T$, Client number $K$, learning rate $\eta$, the unlearning client set $\calC$, dataset $\calD_{k_0}, k_0\in \calC$ including remaining data $\calD_{k_0}^r$ and unlearning data $\calD_{k_0}^u$ for unlearning client $k_0$.
\begin{algorithmic}[1]
  \STATE Initialize the feature extractor $E$, unlearning learning module $W^l$ and auxiliary unlearning module $W_{k_0}^a$
\FOR{$t = 1, 2, \dots, T$} 
    \colorbox{rgb:red!2,65;green!30,60;blue!20,125}{
		\parbox{0.82\textwidth}{\vbox{\STATE \gray{$\triangleright$ \textit{Clients perform:}}
    \FOR{Client $k$ in $\{1,\dots,K\}$}
    \STATE Set $E_k = E$, $W_k^l = W^l$;
    \STATE Compute the learning loss $\tilde{\ell}=\ell(\calD_k; E_k, W_k^l)$;
    \STATE $W_{k}^l \longleftarrow W_k^l - \eta \nabla_{W_k^l}\tilde{\ell}$;
    \STATE $E_k \longleftarrow E_k - \eta \nabla_{E_k}\tilde{\ell}$;
    \ENDFOR }}}
     \colorbox{rgb:red!2,65;green!30,90;blue!20,125}{
		\parbox{0.82\textwidth}{\vbox{
   \FOR{Client $k_0$ in $\calC$}
   \STATE Set $W_{k_0}^a = W^a$;
  \STATE Set $\calD^u_{k_0} = (x_{k_0,i}^u, y^{u'}_{k_0,i} \sim U(1,C))$;
     \STATE Set $\calD^{r'}_{k_0} =\calD^{r}_{k_0}$;
    \STATE Set $\calD^{'}_{k_0} = \calD^{u'}_{k_0} \cup \calD_{k_0}^{r'}$;
    \STATE Compute the learning loss $\tilde{\ell}=\ell(\calD^{'}_{k_0} ; E_{k_0},W_{k_0}^a) $;
    \STATE $W_{k_0}^a \longleftarrow W_{k_0}^a - \eta \nabla{W_{k_0}^a}\tilde{\ell}$;
    \ENDFOR
   }}} 
    \STATE Upload the $W_k^l$ and $E_k$ to the server;
\STATE \gray{$\triangleright$ \textit{The server performs:}}
\STATE The server aggregates $E$ and $W^l$ as: 
\begin{align*}
        W^l = \frac{1}{K} \sum_{k=1}^KW_k^l; E = \frac{1}{K} \sum_{k=1}^KE_k; W^a = \frac{1}{|\calC|}\sum_{k_0\in \calC} W_{k_0}^a
\end{align*}
\STATE The server distributes $E$ and $W^l$ to all clients.
\ENDFOR
\STATE \colorbox{rgb:red!30,155;green!20,20;blue!20,30}{\parbox{0.92\textwidth}{\vbox{The server implements unlearning process: 
\begin{equation*}
    \hat{W} = \alpha W^l + (1-\alpha) W^a 
\end{equation*}}}} \\
  \RETURN $E, \hat{W}$
\end{algorithmic}
\end{algorithm}
\end{minipage}
\hspace{0.4cm}
\begin{minipage}{0.48\textwidth}
\begin{algorithm}[H]
\small
	\caption{Unlearning Class in FL for multiple clients (\colorbox{rgb:red!2,65;green!30,60;blue!20,125}{Learning Module}, \colorbox{rgb:red!2,65;green!30,90;blue!20,125}{Auxiliary Unlearning Module} and \colorbox{rgb:red!30,155;green!20,20;blue!20,30}{Linear Operation})}\label{algo:unlearn-class-mul}
\textbf{Input:} Communication rounds $T$, Client number $K$, learning rate $\eta$, the unlearning client set $\calC$, dataset $\calD_{k_0}, k_0\in \calC$ including remaining data $\calD_{k_0}^r$ and unlearning data $\calD_{k_0}^u$ for unlearning client $k_0$.
\begin{algorithmic}[1]
  \STATE Initialize the feature extractor $E$, unlearning learning module $W^l$ and auxiliary unlearning module $W_{k_0}^a$
\FOR{$t = 1, 2, \dots, T$} 
    \colorbox{rgb:red!2,65;green!30,60;blue!20,125}{
		\parbox{0.82\textwidth}{\vbox{\STATE \gray{$\triangleright$ \textit{Clients perform:}}
    \FOR{Client $k$ in $\{1,\dots,K\}$}
    \STATE Set $E_k = E$, $W_k^l = W^l$;
    \STATE Compute the learning loss $\tilde{\ell}=\ell(\calD_k; E_k, W_k^l)$;
    \STATE $W_{k}^l \longleftarrow W_k^l - \eta \nabla_{W_k^l}\tilde{\ell}$;
    \STATE $E_k \longleftarrow E_k - \eta \nabla_{E_k}\tilde{\ell}$;
    \ENDFOR }}}
     \colorbox{rgb:red!2,65;green!30,90;blue!20,125}{
		\parbox{0.82\textwidth}{\vbox{
 \FOR{Client $k_0$ in $\calC$}
  \STATE Set $W_{k_0}^a = W^a$;
\STATE Set $\calD^{u'}_{k_0} = \calD^u_{k_0}$;
\STATE Set $\calD^{r'}_{k_0} = (x_{k_0,i}^r, c ) $; 
\STATE Set $\calD^{'}_{k_0} = \calD^{u'}_{k_0} \cup \calD_{k_0}^{r'}$;
\STATE Compute the learning loss $\tilde{\ell}=\ell(\calD^{'}_{k_0} ; E_{k_0}, W_{k_0}^a,) $;
\STATE $W_{k_0}^a \longleftarrow W_{k_0}^a - \eta \nabla_{W_{k_0}^a}\tilde{\ell}$;
\ENDFOR
}}}
\STATE Upload the $W_k^l$ and $E_k$ to the server;

\STATE \gray{$\triangleright$ \textit{The server performs:}}
\STATE The server aggregates $E$ and $W^l$ as: 
\begin{equation*}
    W^l = \frac{1}{K} (W_1^l+ \cdots + W_K^l); E = \frac{1}{K} (E_1+ \cdots + E_K)
\end{equation*}
\STATE The server distributes $E$ and $W^l$ to all clients.
\ENDFOR
  \STATE \colorbox{rgb:red!30,155;green!20,20;blue!20,30}{\parbox{0.92\textwidth}{\vbox{The server implements unlearning process: 
  \begin{equation*}
      \hat{W} =  W^l -\beta \sum_{k_0\in \calC}W^a_{k_0} 
  \end{equation*}}}} \\
  \RETURN $E, \hat{W}$
\end{algorithmic}
\end{algorithm}
\end{minipage}
\vspace{-0.2cm}
\end{figure*}


\section{More Experiment}
This section presents additional experiments that further support the conclusions stated in the main text.
In Section \ref{subsec:app1}, we provide a re-demonstration of the advantages of FedAU in terms of efficiency, including time and memory.
Section \ref{subsec:app6} introduces a strategy to reduce the training time of FedAU. In Section \ref{subsec:app2}, we leverage a new metric called attack recall in MIA (Membership Inference Attack) to evaluate the performance of different FMU methods. Section \ref{subsec:app3} compares FedAU to FedRecovery \cite{zhang2023fedrecovery}, another state-of-the-art FMU method. Sections \ref{subsec:app4} and \ref{subsec:app5} provide an analysis of the results obtained on the CIFAR100 dataset. Finally, in Section \ref{subsec:app5}, we conduct ablation studies to investigate the impact of various factors on the performance of FedAU. We explore the position of the auxiliary unlearning module, the coefficients ($\alpha$ and $\beta$), and the proportion of unlearning samples, shedding light on the sensitivity of FedAU to these parameters.

Through these additional experiments and analyses, we aim to further validate the effectiveness, efficiency, and robustness of FedAU, strengthening the conclusions drawn in the main text.
\subsection{Reduced Efficiency}\label{subsec:app1}
For the Retrain scheme, we calculate the total time of 200 rounds of retraining for comparison purposes. For our proposed scheme (FedAU), the Amnesiac unlearning \cite{graves2021amnesiac} and FedRecovery \cite{zhang2023fedrecovery} schemes, we calculate and display the time required to perform the unlearn operation. This helps us understand the time cost associated with unlearning. In particular, we set relearning of the FedEraser \cite{liu2021federaser} as 50 epochs, and set finetuning round of Class-dis \cite{wang2022federated} scheme as 10 rounds after pruning the model channels. We re-demonstrate the unlearning cost in Tab. \ref{tab:eff-app}. It shows: 
\begin{itemize}
    \item Among all the schemes, the Retraining scheme and FedEraser, which involve fine-tuning operations, consume considerably more time compared to other methods;
    \item Although the Amnesiac and FedRecovery schemes require a relatively small amount of time for unlearning, they are still several orders of magnitude slower than FedAU;
    \item In terms of memory cost, FedAU has the smallest memory requirement. On the other hand, the Amnesiac and FedRecovery schemes incur a significant memory overhead since they need to store gradients from each training epoch to facilitate the unlearning process.
\end{itemize} 
\begin{table}[htbp] 
\centering
\small
\caption{Unlearning efficiency for different FMU methods on AlexNet-CIFAR10.}
\label{tab:eff-app}
\begin{tabular}{@{}ccc@{}}
\toprule
FMU methods & Unlearning Time &  Memory \\ \midrule
Retraining    & $\sim 10^3$s &  \textbf{$\sim 10^2$}MB   \\ \midrule
Class-dis \cite{wang2022federated} &  $\sim 10^2$ s   &  \textbf{$\sim 10^2$}MB           \\  \midrule\
Amnesiac \cite{graves2021amnesiac}   & $\sim 10^0$s  &  $\sim 10^4$MB       \\ \midrule
FedEraser \cite{liu2021federaser}  & $\sim 10^3$s  &  \textbf{$\sim 10^2$}MB       \\ \midrule
FedRecovery \cite{zhang2023fedrecovery} & $\sim 10^0$s&  $\sim 10^4$MB  \\  \midrule
FedAU (Ours)       & \textbf{$\sim 10^{-3}$}s &  \textbf{$\sim 10^2$}MB  \\ \bottomrule
\end{tabular} 
\end{table}

\subsection{Training Cost} \label{subsec:app6}
Based on the experimental results presented in Table \ref{tab:fedau-im}, it is evident that the Ul-Acc (Unlearning Accuracy) drops below 2\% within 10 training epochs. This implies that the auxiliary unlearning module can be effectively learned within a relatively small number of training epochs. Therefore, it is not necessary to train the auxiliary unlearning module for an extended period.

\begin{table}[htbp]
\begin{tabular}{@{}l|rrrrrr@{}}
\toprule
Training Epoch & 1     & 2                           & 3                           & 4                           & 5                           & 6                           \\ \midrule
MNIST          & 11.56 & 2.55                        & 2.05                        & 1.44                        & 1.25                        & 1.25                        \\
CIFAR10        & 6.8   & \cellcolor[HTML]{FFFFFF}4.2 & \cellcolor[HTML]{FFFFFF}2.2 & \cellcolor[HTML]{FFFFFF}1.8 & \cellcolor[HTML]{FFFFFF}1.6 & \cellcolor[HTML]{FFFFFF}1.5 \\
CIFAR100       & 4.8   & 3.8                         & 2.6                         & 2                           & 0.8                         & 0.6   \\ \hline
\end{tabular}
\caption{Ul-Acc (\%) with change of Training Epoch for FedAU of the different dataset.}\label{tab:fedau-im}
\end{table}

These findings suggest that a few training epochs are sufficient to achieve satisfactory performance in learning the auxiliary unlearning module. By training the module for a limited number of epochs, we can effectively reduce the time and computational resources required for the unlearning.

\subsection{Evaluation of Different FMU Methods via MIA}\label{subsec:app2}
We utilize the membership inference attack (MIA) by determining whether the unlearning data is the training data of the unlearning model. Moreover, we use the attack recall \cite{graves2021amnesiac} and the low attack recall indicates the good unlearning effect.

Experimental results in Tab. \ref{tab:mia} compare different FMU methods under unlearning class scenarios on CIFAR10. They show FedAU also performs as well as Retraining method, i.e., the attack accuracy of MIA for FedAU is zero.

\begin{table}[htbp]
\centering
\renewcommand\arraystretch{1.6}
\setlength{\tabcolsep}{1.0mm}
\begin{tabular}{@{}llll@{}}
\toprule
           \% & \multicolumn{1}{c}{Retraining} & Amnesiac & \multicolumn{1}{c}{FedAU (Ours)} \\ \hline
Attack Acc & 0.00                             & 12        & 0.00   \\
\bottomrule
\end{tabular}
\caption{Attack accuracy via MIA \protect\cite{graves2021amnesiac} for different FMU methods on CIFAR10 under unlearning class scenario.} \label{tab:mia}
\end{table}

\subsection{FedAU v.s FedRecovery}\label{subsec:app3}
We compare FedAU to FedRecovery \cite{zhang2023fedrecovery} in Tab. \ref{tab:fedre} under unlearning a client. It shows FedAU performs better than FedRecovery with lower Ul-Acc and high Rm-Acc.

\begin{table}[htbp]
\small
\renewcommand\arraystretch{1.6}
\setlength{\tabcolsep}{1.0mm}
\begin{tabular}{@{}c||c|cc|cc|cc}
\toprule
                          \% & FedAvg                        & \multicolumn{2}{c|}{FedAU (ours)}                                                         & \multicolumn{2}{c|}{Retrain}                                                      & \multicolumn{2}{c}{FedRecovery}                                  \\ \hline
                          & Rm-Acc                           & \multicolumn{2}{c|}{Ul/Rm-Acc}                & \multicolumn{2}{c|}{Ul/Rm-Acc}                & \multicolumn{2}{c}{Ul/Rm-Acc}   \\ \cline{2-8}
\multirow{-2}{*}{CIFAR10} & 87.49 & 0.52 & {86.83} &0.00 & {79.09} & 9.48 & \multicolumn{1}{r}{68.75}         \\   \hline                         
\end{tabular}
\caption{Comparison between FedAU and FedRecovery \protect\cite{zhang2023fedrecovery} On AlexNet-CIFAR10 for unlearning client.} \label{tab:fedre}
\end{table}

\begin{table*}[htbp]
\centering
\caption{The comparison with current methods, including FedAvg, Retraining, Amnesiac unlearning \protect\cite{graves2021amnesiac}, Class-dis \protect\cite{wang2022federated} and Federaser \protect\cite{liu2021federaser} and FedAU in different federated machine unlearning scenarios on CIFAR100.}
\label{tab:unlearn_result-cifar100}
\renewcommand\arraystretch{1.6}
\setlength{\tabcolsep}{1.0mm}
\begin{tabular}{@{}c|c||c|cc|cc|cc|cc|cc|@{}}
\toprule
& & \multicolumn{1}{c|}{FedAvg} &  \multicolumn{2}{c|}{Retraining} & \multicolumn{2}{c|}{Amnesiac} & \multicolumn{2}{c|}{Class-disc} & \multicolumn{2}{c|}{FedEraser} & \multicolumn{2}{c|}{FedAU} \\
\cline{3-13}
\multirow{-2}{*}{\begin{tabular}[c]{@{}c@{}}Dataset\\ (\%)\end{tabular}} & \multirow{-2}{*}{\begin{tabular}[c]{@{}c@{}}UL\\ Method\end{tabular}} & Rm-Acc & \multicolumn{2}{c|}{Ul/Rm-Acc} & \multicolumn{2}{c|}{Ul/Rm-Acc} & \multicolumn{2}{c|}{Ul/Rm-Acc} & \multicolumn{2}{c|}{Ul/Rm-Acc} & \multicolumn{2}{c|}{Ul/Rm-Acc} \\
\midrule
& Samples &   56.12 & 0.6& 	54.56 & 0&	53.88& --- & --- & --- & ---  & 3.6 &	55.68\\
& Classes & 55.48 & 0	& 54.44 & 28.1	&43.23 &  0&	48.65 & --- & --- &0	&54.5 \\
\multirow{-3}{*}{\begin{tabular}[c]{@{}c@{}}CIFAR100\\ ResNet\end{tabular}} & Clients & 55.15 &0.1& 	52.77&  0	&52.74 &--- & --- &0.1	& 51.12 &0	&50.22\\
\bottomrule
\end{tabular}
\end{table*}

\subsection{Comparison of Different FMU Methods on CIFAR100} \label{subsec:app4}

We compare different FMU methods on CIFAR100 in Tab. \ref{tab:unlearn_result-cifar100}. It shows FedAU also performs well in CIFAR100 in unlearning samples, class and client. Specifically, the Ul-Acc and Rm-Acc only drops below 3\% and 0.5\% compared to retraining and FedAvg respectively.

\subsection{Ablation Study}\label{subsec:app5}
\subsubsection{Analysis on Position of Auxiliary Unlearning Module} 
We present an analysis investigating the influence of the position of the auxiliary unlearning module $W_{k_0}^a$ within the FedAU framework (we remove the Relu layer of the last three layer of LeNet in convenience). The results are summarized in Tab. \ref{tab:arch}. Notably, our findings reveal that regardless of the placement of within the fully connected layers, both Ul-Acc (Unlearning Accuracy) and Rm-Acc (Remaining Accuracy) exhibit close proximity to the retraining method.
This observation serves as compelling evidence, demonstrating the capability of FedAU to effectively train the auxiliary unlearning module $W_{k_0}^a$ in any fully connected layer.

\begin{table}[htbp]
\centering
\renewcommand\arraystretch{1.2}
\setlength{\tabcolsep}{1.4mm}
\begin{tabular}{@{}l||cc|ll@{}}
\toprule
\multirow{2}{*}{Position of $W_{k_0}^a$} & \multicolumn{2}{c|}{Retrain}                  & \multicolumn{2}{l}{FedAU (Ours)}                      \\ \cline{2-5}
                          & \multicolumn{2}{c|}{Ul/Rm-Acc \%}                & \multicolumn{2}{c}{Ul/Rm-Acc \%}                         \\ \hline
Layer 1                   & \multirow{3}{*}{0.00} & \multirow{3}{*}{99.33} & 0.83	 & 98.36                        \\
Layer 2                   &                    &                         &0.43	 & 98.77                         \\
Layer 3                   &                    &                      & 0.13 & 98.87 \\ \bottomrule
\end{tabular}
\caption{The impact of position of auxiliary unlearning module $W_{k_0}^a$ of FedAU for unlearning 10\% samples on MNIST.} \label{tab:arch}
\end{table}

\subsubsection{Analysis on Coefficient $\alpha$ and $\beta$} 
\begin{figure}[htbp]
    \centering
     \begin{subfigure}{0.22\textwidth}
  		 	  \includegraphics[width = \linewidth]{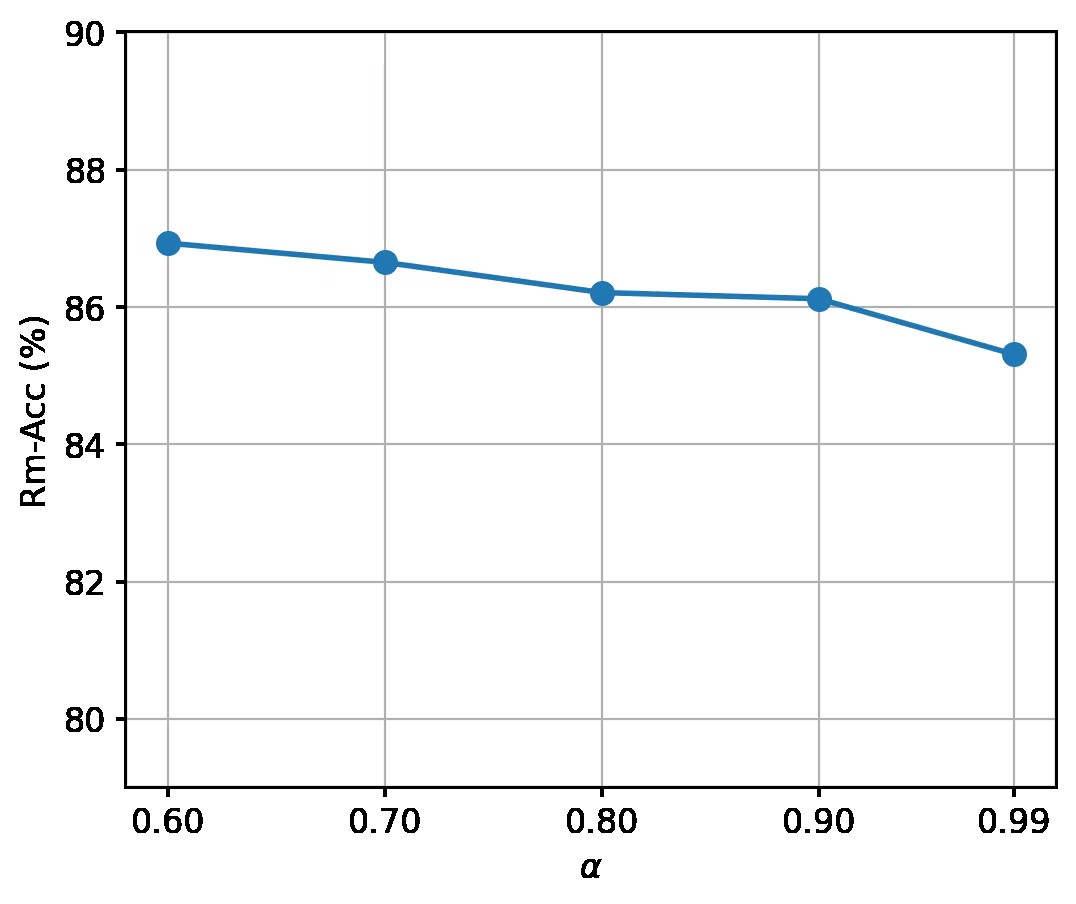}
    	\end{subfigure}
    \begin{subfigure}{0.22\textwidth}
  		 \includegraphics[width=1\textwidth]{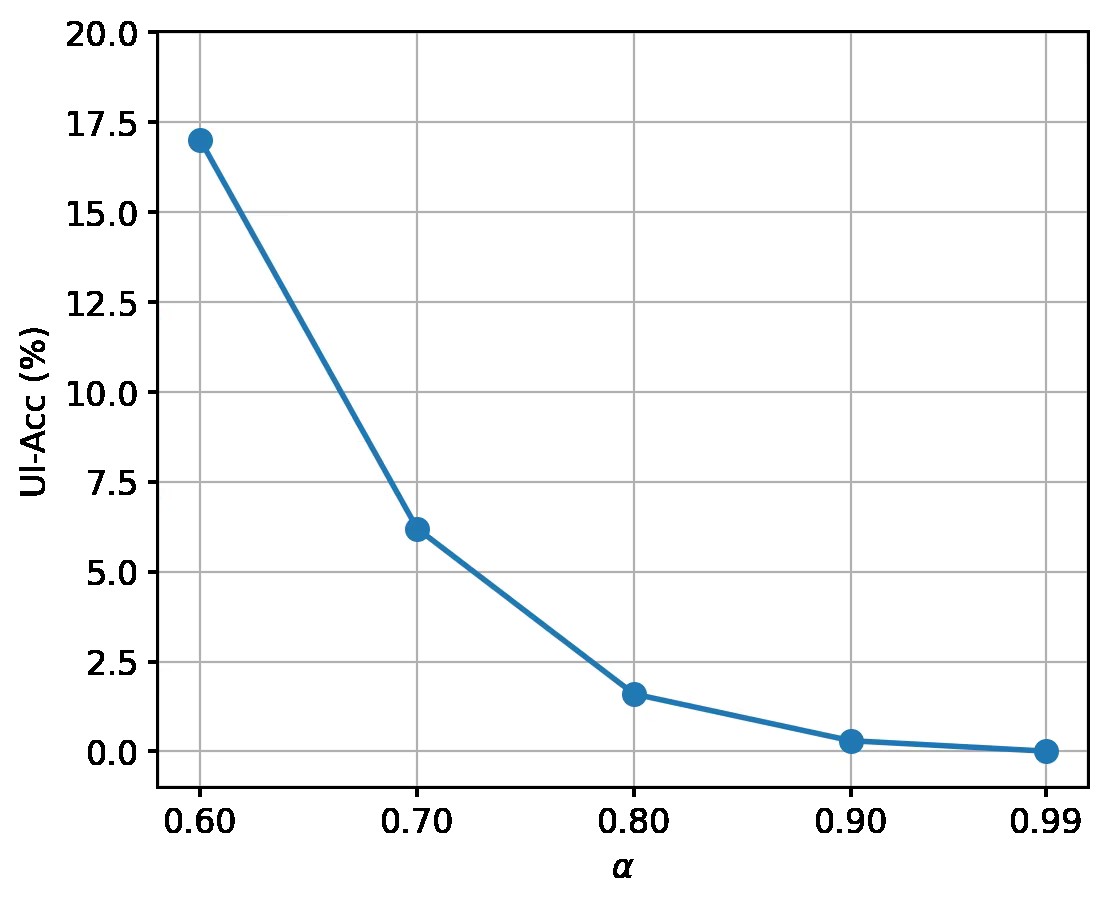}
    		\end{subfigure}

           \begin{subfigure}{0.22\textwidth}
  		 	  \includegraphics[width = \linewidth]{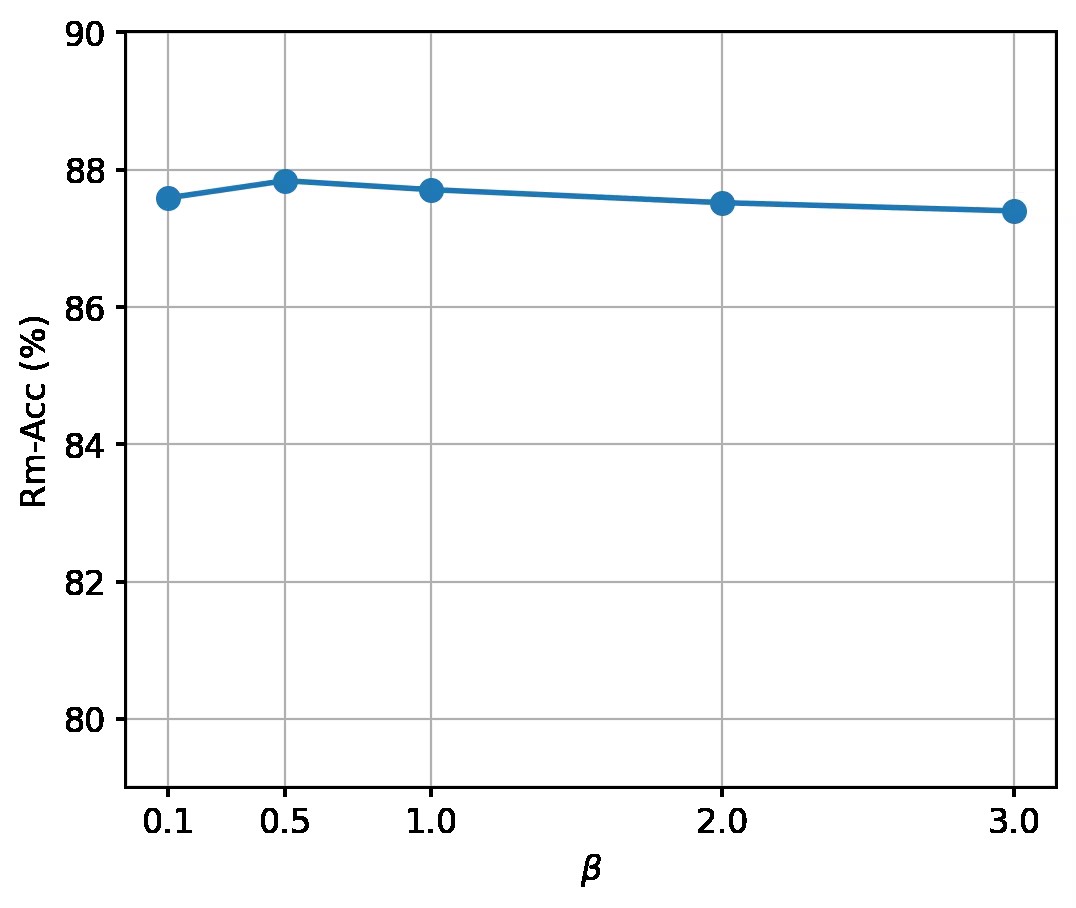}
    	\end{subfigure}
    \begin{subfigure}{0.22\textwidth}
  		 \includegraphics[width=1\textwidth]{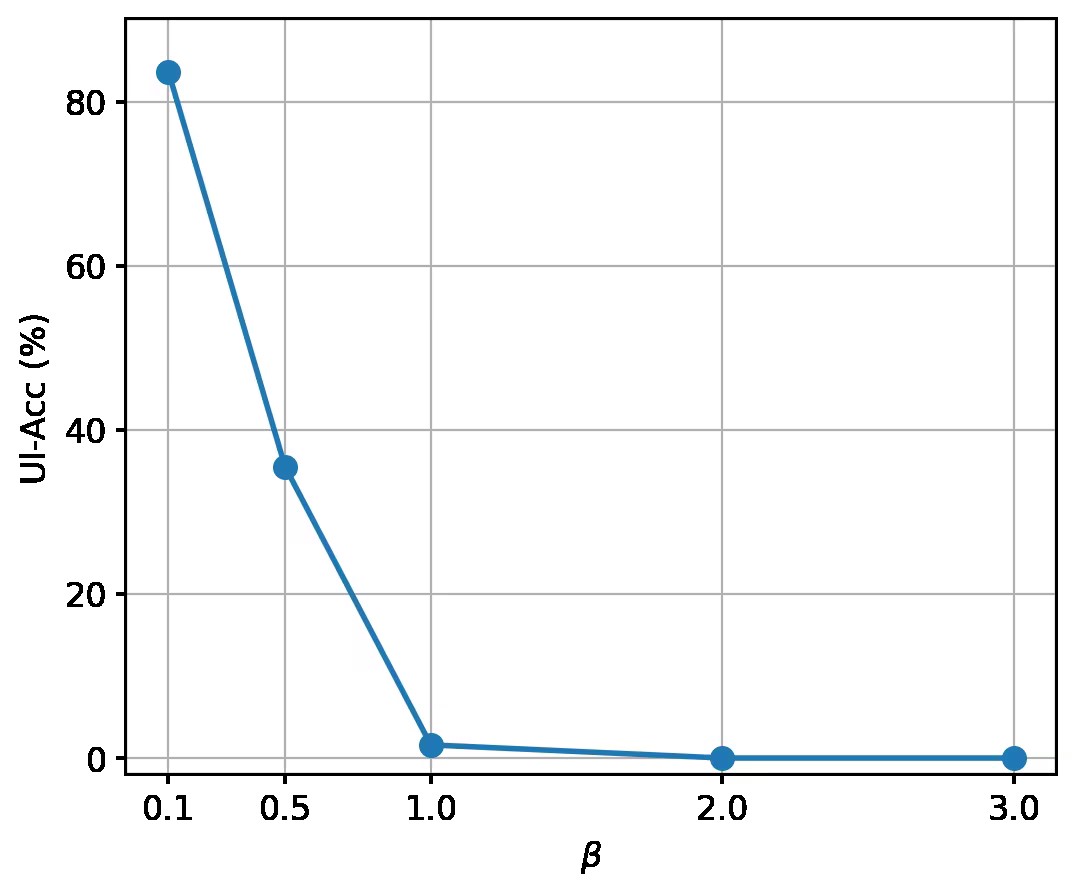}
    		\end{subfigure}
    \caption{The impact of coefficient $\alpha$ and $\beta$ for the proposed FedAU method on CIFAR10. }
  
    \label{fig:coef}
    
\end{figure}

We provide the analysis on the accuracy of unlearning data and remaining data with the change of $\alpha$ and $\beta$. Results in Fig. \ref{fig:coef} illustrate:
\begin{itemize}
    \item When $\alpha$ tends to zero, both the Ul-Acc and Rm-Acc become small indicating the better unlearning effect and the worse model utility. This is because the proportion of private auxiliary unlearning module $W^a_{k_0}$ goes large such that unlearning effect is more obvious but accuracy on the other clients' data decreases.
    \item When $\beta$ goes large, the Ul-Acc becomes small indicating the better unlearning effect. This is because a large $\beta$ can guarantee the largest logit of unlearning class is removed.
\end{itemize}
\subsubsection{Analysis on the Proportion of Unlearning Samples} 
\begin{figure}[htbp]
    \centering
     \begin{subfigure}{0.22\textwidth}
  		 	  \includegraphics[width = \linewidth]{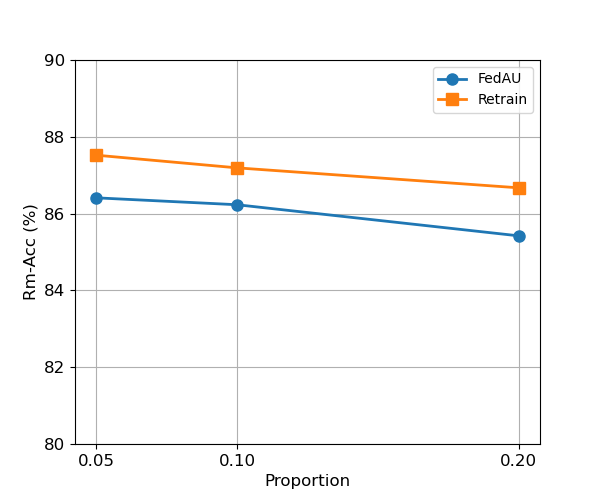}
    	\end{subfigure}
    \begin{subfigure}{0.22\textwidth}
  		 \includegraphics[width=1\textwidth]{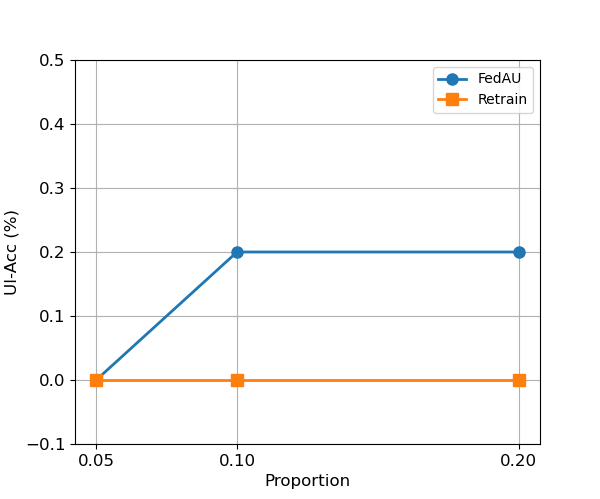}
    		\end{subfigure}

    \caption{The impact of Proportion of Unlearning Samples for the proposed FedAU method on CIFAR10. }
  
    \label{fig:prop}
    
\end{figure}

For Unlearning samples, we analyze the influence on the proportion of unlearning samples for the Ul-Acc and Rm-Acc. We set the unlearning client have the total 5000 training data and aims to forget some proportion of the training data. Results in Fig. \ref{fig:prop} illustrate the Ul-Acc becomes higher and Rm-Acc goes lower with the proportion of unlearning samples increases.

\clearpage
\section{Proof}
This section provides the proof for Proposition 1 and Theorem 1.
\begin{prop}
Consider two fully connected layers projecting the input $x \in \RR^{m_2}$ to the logit $l \in \RR^{m_1}$ as : $l_1= w_1x + b_1, l_2 = w_2x +b_2 $, then the linear operation of weights $w_1, b_1$ and $w_2, b_2$ has the same influence on logits $l_1$ and $l_2$.
\end{prop}
\begin{proof}
Consider the linear operation between $w_1$ and $w_2$ as: $\alpha_1 w_1 +\alpha_2 w_2, \alpha_1 b_1 +\alpha_2w_2$; then
\begin{align}
    &\alpha_1 l_1 + \alpha_2 l_2   \\
    &=\alpha_1(w_1x + b_1) + \alpha_2(w_1x + b_1)\\
    &=(\alpha_1w_1+ \alpha_2w_2) x + (\alpha_1b_1+ \alpha_2b_2)
\end{align}
Therefore, the linear operation of weights $w_1, b_1$ and $w_2, b_2$ has the same influence on logits $l_1$ and $l_2$.
\end{proof}

\begin{theorem} \label{thm:thm1-app}
For client $k_0$ aims to remove $\calD^u_{k_0}$ from the $\calD_{k_0}$, and let $\calD^r_{k_0} = \calD_{k_0} - \calD^u_{k_0} $. There exist $\alpha$ and $\beta$ such that both unlearning Algorithm \ref{algo:unlearn-sample} and \ref{algo:unlearn-class} satisfy the requirement \eqref{eq:rq1} and \eqref{eq:rq2}, i.e., 
\begin{equation} \label{eq:upload-gradients-app}
\left\{
\begin{aligned}
&\argmax_iF^i_{W^l}(x) = \argmax_iF^i_{\hat{W}}(x),\quad x\in \calD^r_{k_0},\\
&\argmax_i F^i_{\hat{W}}(x) \neq y, \qquad \qquad \qquad (x,y)\in \calD^u_{k_0},\\
\end{aligned}
\right.
\end{equation}
\end{theorem}

\begin{proof}
We firstly prove Algo. \ref{algo:unlearn-sample} satisfies the requirement \eqref{eq:rq1} and \eqref{eq:rq2}. Since $W^a_{k_0}$ and $W^l$ are both learnt by the $\calD_{k_0}^r$, then 
\begin{equation} \label{eq:ex1}
    \argmax_iF^i_{W^l}(x) = \argmax_iF^i_{W^a_{k_0}}(x),\quad x\in \calD^r_{k_0}.
\end{equation}
Therefore, for any $x\in \calD^r_{k_0}$, we have
\begin{align}
     &\argmax_iF^i_{\hat{W}}(x) \\
     &= \argmax_iF^i_{\alpha W^l + (1-\alpha)W^a_{k_0}}(x) \\
     &= \argmax_i(\alpha F^i_{W^l}(x) + (1-\alpha)F^i_{W^a_{k_0}}(x)) \\
     &= \alpha \argmax_iF^i_{W^l}(x) + (1-\alpha)\argmax_iF^i_{W^l}(x) \\
     & =  \argmax_iF^i_{W^l}(x),
\end{align}
where the second equality and the third equal are due to Proposition 1 and Eq. \eqref{eq:ex1} respectively. Thus, the Algo. \ref{algo:unlearn-sample} satisfies requirement \ref{eq:rq1}. 

Denote $N_1 = \max_{x \in \calD^r_{k_0}, i\in [C]}|F^i_{W^l}(x)|$ and $\delta$ to be the difference between the largest and the second largest value of $F_{W^l}(x)$. For any $(x,y) \in \calD_{k_0}^u$, the new unlearning dataset $(x', y') \sim U(1,C)$, where $U(1,C)$ represents the discrete uniform distribution on value $1,\cdots, y-1,  y+1, \cdots,C$. Therefore, for any $x\in \calD^u_{k_0}$, if $\alpha <\frac{\delta}{\delta +2N_1}<1$, we can obtain
\begin{align}
         &\argmax_iF^i_{\hat{W}}(x) \\
     &= \argmax_iF^i_{\alpha W^l + (1-\alpha)W^a_{k_0}}(x) \\
     &= \argmax_i(\alpha F^i_{W^l}(x) + (1-\alpha)F^i_{W^a_{k_0}}(x)) \\
     &= y' \neq y,
\end{align}
where the last equality is due to $(1-\alpha)\delta > 2\alpha N_1$, which means even the largest value of $F_{W^l}(x)$ adding the minimum value of $F_{W^a_{k_0}}(x)$ is still larger than the second largest value of $F_{W^l}(x)$ adding the largest value of $F_{W^a_{k_0}}(x)$. Consequently, Algo. \ref{algo:unlearn-sample} satisfies requirement \ref{eq:rq2}.

Secondly, we prove Algo. \ref{algo:unlearn-class} satisfies the requirement \eqref{eq:rq1} and \eqref{eq:rq2}. Since $W^a_{k_0}$ is learnt by the $\calD_{k_0}^{r'}$, whose classes are all label $c$, then 
\begin{equation} \label{eq:ex2}
    \argmax_iF^i_{W^l}(x) = c,\quad x\in \calD^r_{k_0}.
\end{equation}
Denote $N_2 = \max_{x \in \calD^r_{k_0}, i\in [C]}|F^i_{W^a_{k_0}}(x)|$ and $\delta$ to be the difference between the largest and the second largest value of $F_{W^l}(x)$. Suppose $N_2$ to be small enough. Therefore, for any $x\in \calD^r_{k_0}$, if $\beta < \frac{\delta}{2N_2}$, we have
\begin{align}
     &\argmax_iF^i_{\hat{W}}(x) \\
     &= \argmax_iF^i_{W^l - \beta W^a_{k_0}}(x) \\
     &= \argmax_i( F^i_{W^l}(x) - \beta F^i_{W^a_{k_0}}(x)) \\
     & =  \argmax_iF^i_{W^l}(x),
\end{align}
where the second equality is due to Proposition 1 and the third equality is because of  $\delta > 2\beta N_2$, which means even the largest value of $F_{W^l}(x)$ subtracting the minimum value of $F_{W^a_{k_0}}(x)$ is still larger than the second largest value of $F_{W^l}(x)$ subtracting the largest value of $F_{W^a_{k_0}}(x)$. Thus, the Algo. \ref{algo:unlearn-class} satisfies requirement \ref{eq:rq1}.
For any $(x,y) \in \calD_{k_0}^u$, for any $(x,c) \in \calD^u_{k_0}$, exists $\beta>0$, we can obtain
\begin{align}
         &\argmax_iF^i_{\hat{W}}(x) \\
     &= \argmax_iF^i_{ W^l - \beta W^a_{k_0}}(x) \\
     &= \argmax_i( F^i_{W^l}(x) - \beta F^i_{W^a_{k_0}}(x)) \\
     &\neq c,
\end{align}
where the last inequality is since existing a large $\beta$ such that the value of index c is small enough. Consequently, Algo. \ref{algo:unlearn-class} satisfies requirement \ref{eq:rq2}.
\end{proof}

\paragraph{\textbf{Discussion.}} Moreover, in FL, the unlearning model further doesn't influence the model performance of other normal clients' data, i.e., the requirement \ref{eq:rq1} is improved to
\begin{equation}
    \argmax_i F^i_{\hat{W}}(x) \neq y, \qquad (x,y)\in \{\calD^u_{k_0}, \calD_j, j\neq k_0\};
\end{equation}
In order to achieve this requirement, to privately learn the auxiliary unlearning model $W_{k_0}^a$ for unlearning client is not enough since $W_{k_0}^a$ cannot classify data $\calD_j, j\neq k_0$ well. Therefore, we set the initialization of $W_{k_0}^a$ to be the global model $W^l$, which have the ability to classify data $\calD_j, j\neq k_0$. 



\end{document}